\documentclass[review]{elsarticle}
 
\usepackage[table,xcdraw]{xcolor}
\usepackage{lineno,hyperref}
\modulolinenumbers[5]

\usepackage{geometry}
\journal{Journal of Information Sciences}

\usepackage{times}
\usepackage{soul}
\usepackage{url}
\usepackage[utf8]{inputenc}
\usepackage{bbding}
\usepackage{graphicx}
\usepackage{amsmath}
\usepackage{booktabs}
\usepackage{algorithm}
\usepackage{algorithmic}
\usepackage{pifont}
\usepackage{graphicx}
\usepackage{subfigure}
\usepackage{multirow}
\usepackage{amsmath}
\usepackage{amsfonts}
\usepackage{color}
\newcommand{\fm}[1]{\textcolor{black}{#1}}

\begin{document}

\begin{frontmatter}

\title{Rethinking the Defense Against Free-rider Attack From the Perspective of Model Weight Evolving Frequency}


\author[mymainaddress,mysecondaryaddress]{Jinyin Chen}
\cortext[mycorrespondingauthor]{Corresponding author}
\ead{chenjinyin@zjut.edu.cn}

\author[mysecondaryaddress]{Mingjun Li}
\ead{limingjuns@outlook.com}

\author[mysecondaryaddress]{Tao Liu}
\ead{leonliu022@163.com}

\author[mymainaddress,mysecondaryaddress]{Haibin Zheng\corref{mycorrespondingauthor}}
\ead{haibinzheng320@gmail.com}

\author[mythirdaryaddress]{Yao Cheng}
\ead{c.candyao@gmail.com}

\author[myfourthaddress]{Changting Lin}
\ead{linchangting@gmail.com}

\address[mymainaddress]{Institute of Cyberspace Security, Zhejiang University of Technology, Hangzhou 310023, China}

\address[mysecondaryaddress]{College of Information Engineering, Zhejiang University of Technology, Hangzhou 310023, China}

\address[mythirdaryaddress]{Huawei International Pte Ltd, Singapore 138589, Singapore}

\address[myfourthaddress]{Binjiang Institute, Zhejiang University, Hangzhou 310014, China}

\begin{abstract}
Federated learning (FL) is a distributed machine learning approach where multiple clients collaboratively train a joint model without exchanging their own data.
Despite FL's unprecedented success in data privacy-preserving, its vulnerability to free-rider attacks has attracted increasing attention.
Numerous defense methods have been proposed for FL to defend against free-rider attacks.
However,
they may fail against highly camouflaged free-riders. And, their defensive effectiveness may sharply degrade when more than 20\% of clients are free-riders.
To address these challenges,
we reconsider the defense from a novel perspective,
i.e., model weight evolving frequency.
Empirically, we gain a novel insight that during the FL's training, the model weight evolving frequency of free-riders and that of benign clients are significantly different.
Inspired by this insight, we propose a novel defense method based on the model \emph{\underline{W}eight \underline{E}volving \underline{F}requency}, referred to as WEF-Defense.
Specifically, we first collect the weight evolving frequency (defined as WEF-Matrix) during local training.
For each client, it uploads the local model's WEF-Matrix to the server together with its model weight for each iteration.
The server then separates free-riders from benign clients based on the difference in the WEF-Matrix.
Finally, the server
uses a personalized approach to provide different global models for corresponding clients,
which prevents free-riders from gaining high-quality models. 
Comprehensive experiments conducted on five datasets and five models demonstrate that WEF-Defense achieves better defense effectiveness ($\sim\!\!\times 1.4$) than the state-of-the-art baselines and
identifies free-riders at an earlier stage of training.
Besides, we verify the effectiveness of WEF-Defense against an adaptive attack and visualize the WEF-Matrix during the training to interpret its effectiveness.
The data and code of WEF-Defense are available at: \url{https://github.com/research-limingjun/WEF-Defense.git}.

\end{abstract}


\end{frontmatter}


\section{Introduction}\label{sec1}


Federated learning (FL) \cite{mcmahan2017communication,mcmahan2016federated,yang2019federated,Justicia21,fd,journals/isci/JiangXZ21,WangZLZL21}, one type of distributed machine learning \cite{books/sp/JiangCZ22,phd/basesearch/Chen22}, has been proposed to train a global model, where clients update the local model parameters, such as the gradients, to the sever without sharing their private data.
Considering the significant advantage in privacy-preserving,
FL has been applied to various data-sensitive practical applications, e.g.,
loan prediction~\cite{long2020federated,conf/icdm/Shingi20},
health assessment~\cite{xu2021federated,KuoP22} and
next-word prediction~\cite{hard2018federated,yang2018applied}.


In a traditional FL system, each client is supposed to contribute its own data for global model training. As a reward, the client has the privilege to use the final trained global model.
In another word, the server usually distributes the final trained global model to each client,
regardless of their contribution.
It leads to the free-rider attack~\cite{lin2019free,journals/corr/abs-2201-09967},
where the clients without any contribution can obtain the high-quality global model.
These clients are called free-riders.
In general, the free-rider issue always exists in a shared resource environment that the free-rider who enjoys the benefits from the environment without any contribution. This issue is well studied in several areas, e.g., stock market~\cite{journals/isci/LiuM22}, transport~\cite{journals/jcmse/JiangXFSL21}, distributed system~\cite{journals/soco/LiHWZLLCL22}, etc.


In this paper, we study the free-rider attack in the FL system.
Noting that several existing works are presented to address free-rider issue in FL,
mainly including two aspects,
outlier detection~\cite{lin2019free,zong2018deep,JiangLDS16} of model parameters and clients' contribution evaluation~\cite{fair1,fair2}.
STD-DAGMM~\cite{lin2019free} is a typical outlier detection method.
It is deployed on the server through a deep autoencoder Gaussian mixture model,
which can detect free-riders as outliers through the learned features of model update parameters.
However, it requires enough benign clients to pre-train the autoencoder.
Additionally, the model updates are easy to disguise for free-riders.
Notably, it is difficult to distinguish the free-riders from benign clients once the number of free-riders are exceed $20\%$.
CFFL~\cite{lyu2020collaborative} is a defense approach that which the server evaluates the contribution of each client based on the validation dataset.
However, there is a strong assumption that the server has enough validation data in real-world FL scenarios.
Advanced free-rider can adopt camouflage that has little effect on the model accuracy, hence its contribution will not decrease rapidly. As a result, the free-rider can obtain the global model rendering the defense invalid.
RFFL~\cite{xu2021reputation} proposes that the server evaluates each client's contribution based on the cosine similarity of the global model gradient and the local model gradient, which may be less effective when clients' data are non-independent and identically distributed (Non-IID)~\cite{noniid1,journals/ijon/ZhuXLJ21}.

The existing defense methods against free-rider attacks are still challenged in three aspects, i.e., 1) to defend against advanced camouflaged free-riders, 2) to tackle in the scenario where multiple free-riders exist (more than 50\% of clients), and 3) to balance the main task performance and the defense effect.
To overcome these challenges, we reconsider the difference between benign clients and free-riders during the dynamic training process.
Surprisingly, we observe that free-riders are able to use the global model that is aggregated and distributed by the server to disguise model weights similar to benign clients,
but are unable to disguise the process of model weights optimization.
The reason is that free-riders do not perform normal training,
thus they cannot evolve as efficiently as benign clients.
Therefore, we intuitively consider leveraging the model evolving information to identify free-riders.
We define the evolving frequency of model weights,
a statistic value that does not involve private information,
to measure the difference between free-riders and benign clients,
which records model weights with drastically varying values.



\begin{figure}[thb]\label{fig2}
\centering{
        \includegraphics[width=1\textwidth,height=0.55\textwidth]{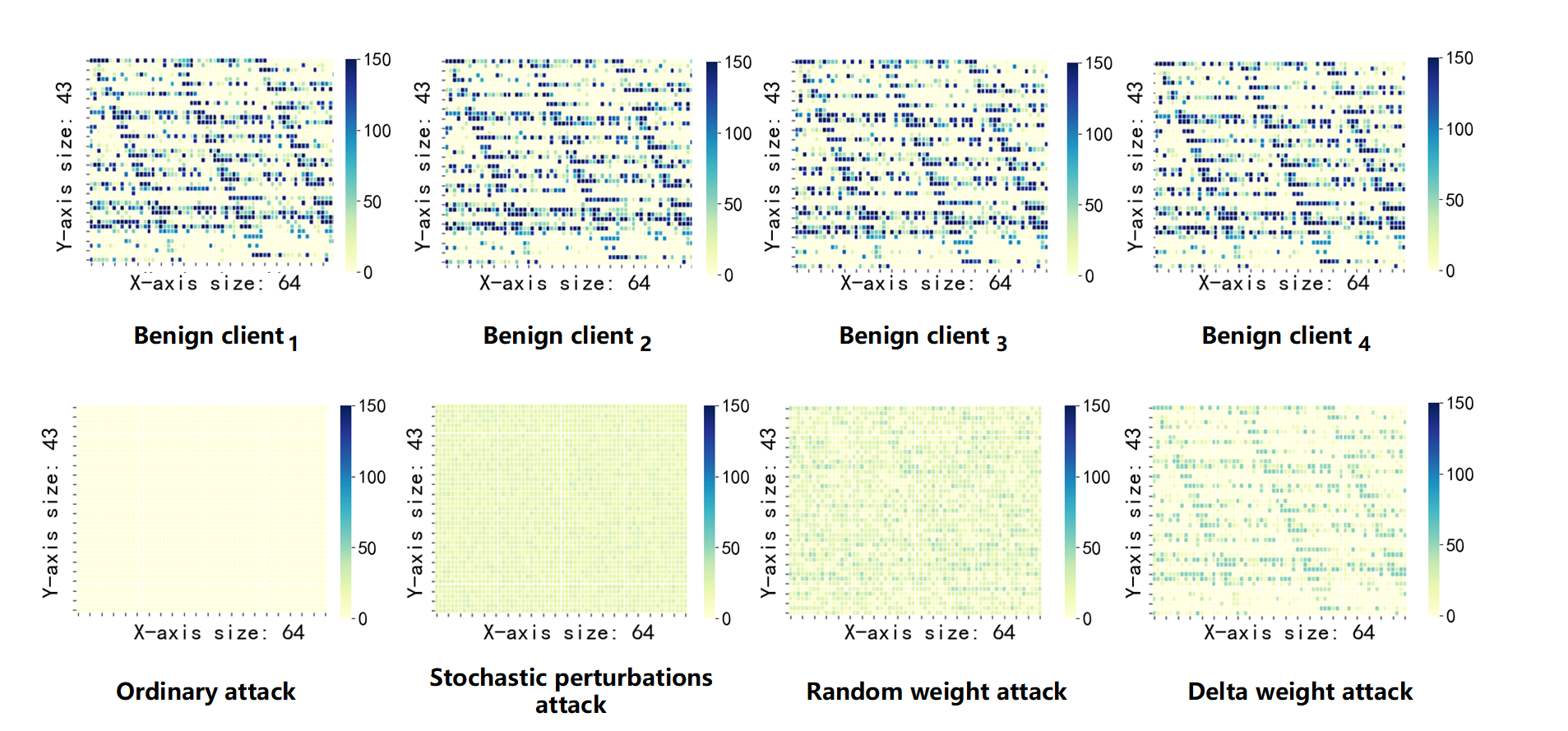} }
\caption{The visualization of the weight evolving frequency for benign clients and free-riders.
We define the concept of weight evolving frequency matrix (WEF-Matrix).
The matrix size is the weight size of the penultimate layer.
Here we use the ADULT dataset and the MLP model as an example,
and the weight size is 86x32.
Considering the aesthetics of the visualization, we adjust the size of the matrix to 43x64.
Each pixel in the figure represents the corresponding weight frequency.
For the evolving frequency of weight,
if the evolution is larger than a calculated threshold,
it is increased by one,
otherwise, it remains unchanged.
}
\label{fig2}
\end{figure}

We visualize the clients' weight evolving frequency in the following example for illustration purposes.
Here is an FL example of training the MLP model~\cite{karlik2011performance} with two fully connected layers and one softmax output layer on the ADULT dataset~\cite{kohavi1996scaling}.
In this example,
there are five clients including four benign clients and one free-rider.
The free-rider executes
ordinary attack~\cite{lin2019free},
stochastic perturbations attack~\cite{fraboni2021free},
random weight attack~\cite{lin2019free} and
delta weight attack~\cite{lin2019free}, respectively.
We visualize the clients' weight evolving frequencies as shown in Fig.\ref{fig2}.
We can observe that during the training process, the weight evolving frequencies of different benign clients are similar, while there is a significant difference between the free-riders and the benign clients,
especially for an ordinary attack,
stochastic perturbations attack and
random weight attack.
Although the weight evolving frequencies of the delta weight attacks are similar to that of the benign clients,
it is worth noting that the scales are different.

Inspired by the difference we observed between the free-riders and the benign clients during the FL training process, we propose a defense method based on \emph{\underline{W}eight \underline{E}volving \underline{F}requency},
referred to as WEF-Defense. Specifically, we define the concept of weight evolving frequency matrix (WEF-Matrix),
to record the weight evolving frequency of the penultimate layer of the model. WEF-Defense calculates the variation of the weight between continuous two rounds of local training,
and takes the average value of the overall variation range as the dynamic threshold to evaluate the evolving frequency of all weights.
Each client needs to upload the local model's WEF-Matrix to the server together with its model weights.
Then, the server can distinguish free-riders from benign clients based on the Euclidean distance,
cosine similarity and overall average frequency of the WEF-Matrix among clients.
For benign clients and free-riders,
the server aggregates and distributes different global models only based on their evolving frequency differences.
In this way,
the global model obtained by free-riders does not have model weights contributed by the benign clients, thereby preventing free-riders from stealing the trained high-quality model.

The main contributions of this paper are summarized as follows.
\begin{itemize}
\item
We first observe that the dynamic information during the FL's local training is different between benign clients and free-riders. We highlight the potential of using the model weight evolving frequency during training to detect free-riders.


\item
Inspired by the observation, we propose WEF-Defense. We design WEF-Matrix to collect the model weight evolving frequency during each client training process and use it as an effective means of detecting free-riders.


\item
Addressing the free-rider attack when major clients are free-riders, i.e., 50\% or even up to 90\%, WEF-Defense adopts a personalized model aggregation strategy~\cite{personalized} to defend the attack in an early training stage.

\item
Extensive experiments on five datasets and five models have been conducted. The results show
that WEF-Defense achieves better defense effectiveness ($\sim\!\!\times 1.4$) than the state-of-the-art (SOTA) baselines and identifies free-riders at an earlier stage of training.
Besides, it is also effective against an adaptive attack. We further provide weight visualizations to interpret its effectiveness.
\end{itemize}

The rest of the paper is organized as follows. Related works are discussed in Section \ref{related}.
The preliminaries, problem statement and methodology are detailed in Sections~\ref{pre}, \ref{problemDef} and \ref{method}, respectively.
The experimental setup and analysis are presented in Sections~\ref{setting} and~\ref{eval}, respectively.
Finally, we discuss our limitation in Section \ref{disccusion} and conclude our work in Section \ref{conclusion}.


\section{Related Work}\label{sec2}\label{related}
In this section, we review the related work and briefly summarize attack and defense methods used as baselines in the experiments.

\subsection{Free-Rider Attacks on Federated Learning}\label{subsec2}
According to the attacker's camouflage tactics,
free-rider attack includes
ordinary attacks~\cite{lin2019free},
random weight attack~\cite{lin2019free},
stochastic perturbations attack~\cite{fraboni2021free} and
delta weight attack~\cite{lin2019free}.

Ordinary attack~\cite{lin2019free} is a primitive attack without camouflage,
where the malicious client does not have any local data,
i.e., it does not perform local training.
By participating in FL training, it obtains the global model issued by the server. 
Based on it, random weight attack~\cite{lin2019free} builds a gradient update matrix by randomly sampling each value from a uniform distribution in a given range $[-R,R]$.
However, it only works well in the condition of an ideal selection of the range value $R$ in advance.
Besides, the randomly generated weight can not generally promise good attack performance by imitating the benign clients' model weights.
Stochastic perturbations attack~\cite{fraboni2021free} is a covert free-rider attack that uploads crafted model weights by adding specific noises to the distributed global model.
In this way, it is difficult for the server to effectively detect the free-riders.
Compared with previous attacks,
delta weight attack~\cite{lin2019free} submits a crafted update to the server by calculating the difference between the last two rounds it received.
Note that for machine learning training, except for the first few epochs, the weight variations at each round are small~\cite{}.
Therefore the crafted updates could be similar to the updates of the benign clients.

\subsection{Defenses against Free-Rider Attacks}
The existing defense methods can be mainly categorized into two
types, i.e., outlier detection of model parameters and clients’ contribution evaluation.

In the first work on the free-rider attack on FL, Jierui et al.~\cite{lin2019free} explored a possible defense based on outlier detection, named STD-DAGMM. Accordingly, the standard deviation indicator is added on the basis of the deep autoencoding Gaussian mixture model~\cite{zong2018deep}. Its network structure is divided into two parts: the compression network and the estimation network.  Specifically, the gradient update matrix is fed into the compression network to obtain the low-dimensional output vector and
the standard deviation from the input vector is calculated, which is then vector superposed with the calculated Euclidean and cosine distance metrics. Finally, concatenate this vector with the low-dimensional representation vector learned by the compression network.
The output concatenated vector is fed into the estimation network for multivariate Gaussian estimation.
However, the time complexity of STD-DAGMM is large,
because each client is required to pre-train its network structure in an early stage. Meanwhile,
when the free-riders take up more than 20\% of total clients,
it is difficult to select a proper threshold to distinguish the free-riders from the benign clients.

The other defense against free-rider attacks is based on clients’ contribution evaluation. Lyu et al.~\cite{lyu2020collaborative} proposed a collaborative fair federated learning, CFFL, to achieve cooperative fairness through reputation mechanisms. It mainly evaluates the contribution of each client using the server's verification dataset. The clients iteratively update their respective reputations, and the server assigns models of different qualities according to their contributions.
The higher the reputation of the clients, the better the quality of the aggregation model obtained. However, CFFL relies on proxy datasets, which is not practical in real-world applications.
On this basis, Xinyi et al.~\cite{xu2021reputation} proposed robust and fair federated learning, RFFL, to realize both collaborative fairness and adversarial robustness through a reputation mechanism.
The server in RFFL iteratively evaluates the contribution of each client by the cosine similarity between the uploaded local gradient and the aggregated global gradient.
Compared with CFFL, RFFL does not require a validation dataset in advance.
However, RFFL is not effective when facing an adaptive free-rider with the ability to camouflage gradients under the Non-IID data.

\section{Preliminaries and Background}\label{sec3}\label{pre}

\subsection{Horizontal Federated Learning}
Compared with the standard centralized learning paradigm,
FL~\cite{fair1,journals/corr/abs-2112-01039} provides a simple and effective solution to prevent private local data from being leaked.
Only global model parameters and local model parameters are allowed to communicate between the server and clients.
All private training data are on the client device, inaccessible to other clients.

As one of the most wildly used FL frameworks,
horizontal FL (HFL) represents a scenario where the training data of participating clients have the same feature space, but have different sample space.
Its training objective can be summed up as searching for the optimal global model:
\begin{equation}
min \left [F \left ( w \right )  := \frac{1}{K}\sum_{i=1}^Kf_i \left (w \right )\right ]
\end{equation}
where $K$ is the number of participating clients,
$f_i$ represents the local model.
Each local model $f_i$ is defined as $f_i  (w ) = L (D_i;w_i)$,
where $D_i$ represents each data sample and its corresponding label,
and $L$ represents the prediction loss using the local parameter $w_i$.

HFL performs distributed training by combining multiple clients, and uses the HFL classic algorithm FedAvg~\cite{mcmahan2017communication}  to calculate the weighted average to update the global model weights $w^{t+1}_g$ as:

\begin{equation}
w_{g}^{t+1}= \frac{1}{K}\sum_{i=1}^Kw^{t+1}_i
\end{equation}
where $t$ is the communication round,
$w^{t+1}_i$ represents the model weights uploaded by the $i$-th client participating in the $(t\!+\!1)$-th round of training.

\subsection{Free-Rider Attack}
Almost all existing free-rider attacks are conducted on the HFL framework,
thus we mainly address the issue of defending against free-riders on HFL.
Free-riders are those clients who have no local data for normal model training,
but aim to obtain the final aggregated model without any contribution.
Since they are involved in the FL process,
free-riders can use some knowledge about the global model 
(e.g., global model architecture, global model weights received at each round)
to generate fake model updates to bypass the server.

\begin{figure}[htb]
\centering{
        \includegraphics[width=0.75\textwidth,height=0.55\textwidth]{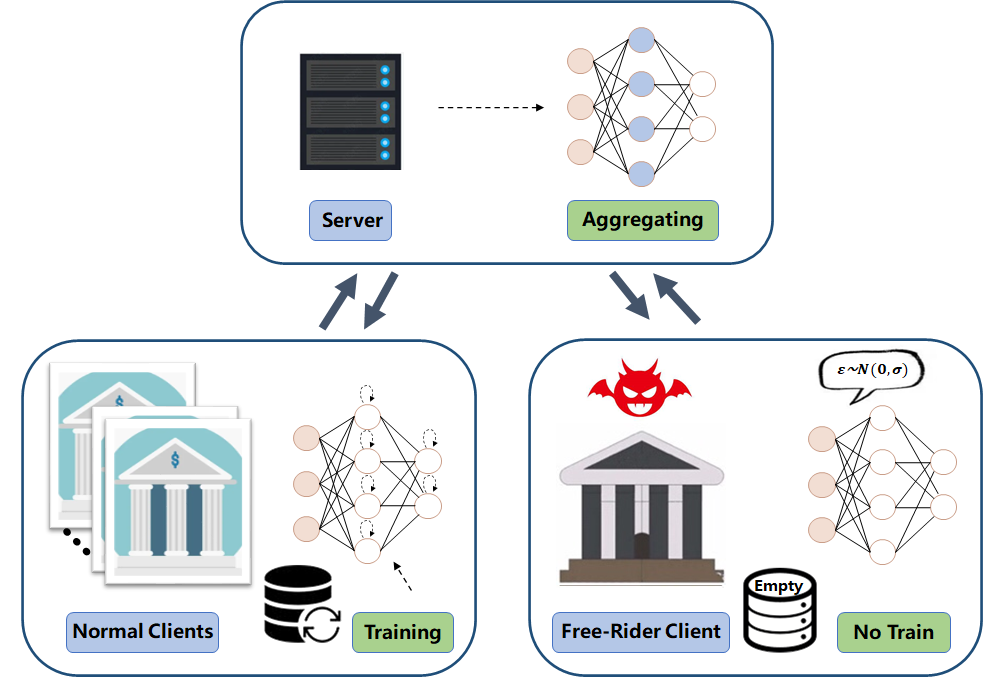} }
\caption{Illustration of a free-rider attack.
The free-rider does not perform normal training,
but transmits fake model updates to the server by adding opportune stochastic perturbations $\varepsilon$ based on Gaussian noise $N(0,\sigma)$. Finally, the global model issued by the server will be distributed to the free-rider.
}
\label{fig1}
\end{figure}

Fig.\ref{fig1} illustrates an example of the free-rider attack in a practical scenario in the financial field, e.g., FL is adopted for the bank's loan evaluation system.
A malicious client may pretend to participate in federated training while concealing the fact that there are no data contributed locally through uploading fake model updates to the server.
Consequently, the free-rider obtains a high-quality model benefiting from other clients' valuable data and computation power.

\section{Problem Statement}\label{problemDef}

\subsection{Problem Formulation}

Suppose there are $K$ clients, denoted by $P=\{p_1$,...,$p_K\}$. The benign clients $P_n$ have a local dataset $D_n$, while the free-riders $P_r$ have no local dataset. Our goal is that in the case of free-riders in the federal system, the central server can distinguish the free-riders from the benign clients to prevent free-riders from stealing a high-quality global model.

\subsection{Assumptions and Threat Model}\label{threatMod}
\textbf{Attacker’s Goal}.
The purpose of a free-rider attack is not to harm the server, but to obtain the global model issued by the server from the federated training without any local data actually contributing to the server.
A free-rider can send arbitrary crafted local model updates to the server in each round of the FL training process, which can disguise itself as a benign client.
The uploaded fake updates have little impact on the performance of the aggregated model, so a high-quality model can be finally obtained by the free-rider.

\textbf{Attacker’s Capability}. We assume that the server is honest and does not know how many free-riders exist among the clients.
If there are multiple free-riders, they can communicate and collude with each other and manipulate their model updates,
but cannot access or manipulate other benign clients' data.
The free-riders have the generally accessible information in an FL system,
including the local model, loss function, learning rate and FL’s aggregation rules.
Free-riders use this knowledge to generate fake model weights $w^{t+1}_r$ to bypass the server. In the $(t\!+\!1)$-th round,
the attack target of free-riders $P_r$ is:

\begin{equation}
w_r^{t+1}=arg~\max_{w_r} \left (C\left (w_g^{t},\psi  \right )  \right )
\end{equation}
where, the camouflage function $C\left ( \cdot  \right )$ uses a set of parameters $\psi$ to process the global model weights issued by the server in the $(t\!+\!1)$-th round, and runs the camouflage method to generate crafted model weights $w_r^{t+1}$ aiming to bypass the free-rider detection and defense methods on the server.
In addition, free-riders can also perform adaptive attacks against specific defense methods, which we discuss in Section \ref{rq5}.

\textbf{Defender’s Knowledge and Capability}.
The server can set up defense methods against free-riders.
But it does not have access to the client's local training data,
nor does it know how many free-riders exist in the FL system.
However, in each training round,
the server has full access to the global model as well as local model updates from all clients.
Additionally, the server can request each local client to upload other non-private information,
and use the information to further defend against free-riders.
The goal of defense can be defined as:

\begin{equation}\label{defense}
w_{g}^{t+1}= \frac{1}{K}\sum_{i=1}^K w^{t+1}_i ,~~~~arg~\max_{w_i}\left (  S( w^{t+1}_i, w^{t+1}_n \right ) )
\end{equation}
where the selection function $S(\cdot)$ selects model updates uploaded by benign clients as much as possible when the model is aggregated.
$w_n^{t+1}$ represents the model weights uploaded by the benign clients,
and $w_i^{t+1}$ represents the selected model weights.
$K$ is the total number of clients.

\section{WEF-Defense}\label{sec4}\label{method}




\subsection{Overview}\label{subsec4}
The concept of sensitive neurons has been widely discussed recently~\cite{XuPZ20,MalmiercaNNPPE19}.
It is observed that when data is input to a neural network, not all neurons will be activated.
Different neurons in different layers will respond to different data features with various intensities, and then the weights will vary significantly.
Free-riders do not have data, and thus they do not have information to take the influence of sensitive and insensitive neurons on parameters into account when they craft their fake updates.
Thus, it is difficult for a free-rider to camouflage the frequency of weight variation. Motivated by it,
WEF-Defense adopts the weight evolving frequency during the local model training process as an effective means to defend against free-rider.
The overview of WEF-Defense is shown in Fig.\ref{fig3},
including three main components:
\textcircled{1} WEF-Matrix information collection (Section \ref{ic}),
\textcircled{2} client separation (Section \ref{sc}),
\textcircled{3} personalized model aggregation (Section \ref{ma}).

\begin{figure}[h]\label{fig3}
\centering{
        \includegraphics[width=1\textwidth]{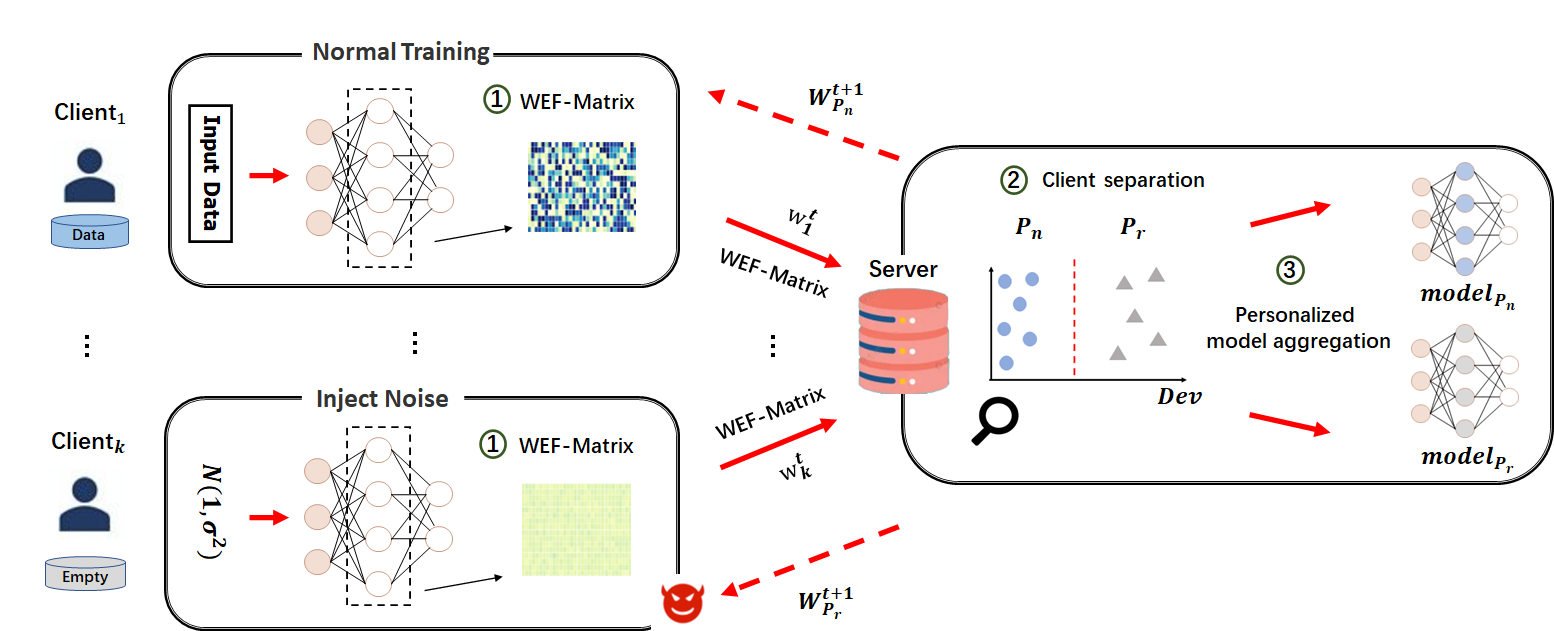} }

\caption{Overview of WEF-Defense. The client uses the initialized WEF-Matrix to record the evolution frequency of the selected layer weight in the local training epoch.
WEF-Matrix is then sent to the server along with the model updates.
According to the WEF-Matrix, the server separates benign clients and free-riders to form  $\left \{P_n,P_r  \right \}$.
The model updates uploaded by the two sets of clients are aggregated separately and delivered to the clients in respective sets only.
}
\label{fig3}
\end{figure}



\subsection{WEF-Matrix Information Collection}\label{ic}
To obtain effective information about the clients,
the WEF-Matrix collection is divided into three steps:
(i) WEF-Matrix initialization,
(ii) threshold determination,
(iii) WEF-Matrix calculation.

\subsubsection{WEF-Matrix Initialization}
We first define the WEF-Matrix,
which is determined by the weights $w_{i,s}\in\mathbb{R}^{H\times W}$ in the penultimate layer of the client $p_i$ and initialized to an all-zero matrix.
It records the information on weight evolving frequency in local training.
We use the weights of the penultimate layer for the following reasons.
The softmax output in the last layer realizes the final classification result.
The closer the weights to the last layer,
the greater they have the impact on the final classification result,
and the more representative the weight variations in this layer are.
The initialization process is as follows:

\begin{equation}\label{equ:initialization}
F^0_i = zeros(H,W) 
\end{equation}
where $zeros(H,W)$ returns an all-zero matrix of size $H\!\times\!W$.
$F_{i}^{0}$ has the same size as $w_{i,s}$.


\subsubsection{Threshold Determination}
We collect the corresponding weight evolving frequency during the local training process of the client through the initialized WEF-Matrix.
Before computing the WEF-Matrix,
we need to determine a dynamic threshold for measuring frequency variations.
Suppose a client $p_i$ is performing the round $(t'+1)$-th local training,
and its model weights obtained after training are $w_i^{t'+1}$.
We select the weights of the client $p_{i}$ in the penultimate layer,
represented as $w_{i,s}^{t'+1}$.
Then, we calculate the weight variations between $w^{t'+1}_{i,s}$ of the $(t'+1)$-th round and $w_{i,s}^{t'}$ of the $t'$-th round,
and take the overall average variation as the threshold.
Calculate the threshold of client $p_{i}$ at the ($t'\!+\!1$)-th round as follows:

\begin{equation}\label{threshold_a}
\alpha_i^{t'+1}= \frac{\sum_{j=1}^{H}\sum_{k=1}^{W}| w^{t'+1}_{i,s,j,k} - w^{t'}_{i,s,j,k} | }{H\times W}
\end{equation}
where $|\cdot |$ returns the absolute value,
$w^{t'+1}_{i,s,j,k}$ is a weight value of the $j$-th row and the $k$-th column from the penultimate layer of the client $p_i$ in the ($t'\!+\!1$)-th round,
$H$ and $W$ represent the rows and columns of $w_{i,s}^{t'+1}$, respectively.


\begin{figure}[h]\label{fig4}
\centering{
        \includegraphics[width=0.5\textwidth,height=0.4\textwidth]{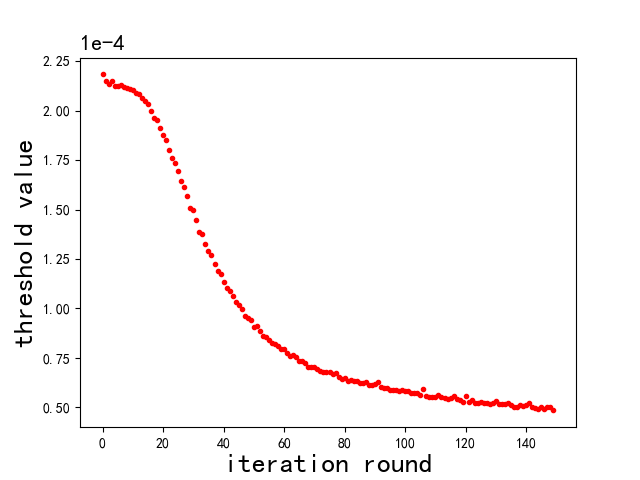} }
\caption{The $\alpha_i$ values for a benign client $p_i$ training MLP model on the ADULT data.}
\label{fig4}
\end{figure}
To find out the evolution of the threshold value during training,
we conduct an experiment to visualize the threshold $\alpha_{i}$ of the $i$-th client during training, shown in Fig.\ref{fig4}.
We use the ADULT data~\cite{kohavi1996scaling} and the MLP model~\cite{karlik2011performance} for illustration.
There are 50 rounds of global training and 3 rounds of local training,
thus in total 150 rounds of iterations.
For a benign client ${p_i}$,
we find that when the model has not converged in the first 60 rounds, the threshold variations greatly. After the model has converged, the threshold fluctuation tends to stabilize.
It illustrates that the $\alpha_i$ is dynamically changed in most training rounds, and this characteristic is difficult to be simulated.

\subsubsection{WEF-Matrix Calculation}
We calculate the weight evolving frequency in local training based on the calculated dynamic threshold. Its calculation process is as follows:

\begin{equation}\label{th}
F_{i,j,k}^{t'+1}= \begin{cases}
  &  F_{i,j,k}^{t'} + 1, ~~~~~~  |w_{i,s,j,k}^{t'+1} -w_{i,s,j,k}^{t'}| > {\alpha}_i^{t'+1}    \\
  & F_{i,j,k}^{t'}      ~~~~~~~,~~~~~~  otherwise
\end{cases}
\end{equation}
where $F_{i,j,k}^{t'+1}$ represents a frequency value of the $j$-th row and the $k$-th column of the client $p_i$ in the $(t'\!+\!1)$-th round,
$j = \left \{1,2,...,H  \right \}$, $k = \left \{1,2,...,W  \right \}$.
The number of frequencies calculated in each round will be accumulated.
Finally, the client uploads the WEF-Matrix together with the model updates to the server.
It is worth noting that the uploaded information does not involve the client's data privacy.

\subsection{Client Separation}\label{sc}
To distinguish benign clients and free-riders, we use the difference of WEF-Matrix to calculate three metrics and combine them to detect free-riders.
The server randomly selects a client $p_i$,
then based on its uploaded WEF-Matrix, calculates
1) the Euclidean distance $Dis$,
2) the cosine similarity $Cos$ with other clients' WEF-Matrix, and
3) the average frequency $Avg$ of their WEF-Matrix,
as follows:
\begin{equation}\label{q1}
Dis_i=\sqrt{\sum_{j\in K}(F_i-F_j)^2  }~~,~i\neq j
\end{equation}
where $F_i$ represents the WEF-Matrix uploaded by the client $p_i$,
$K$ represents the total number of clients.

\begin{equation}\label{q2}
Cos_i=\frac{F_i \cdot F_j}{\|F_i\|\|F_j\|}
\end{equation}
where $\cdot$ represents the matrix dot product, and $||\cdot ||$ represents the 2-norm of the matrix.

\begin{equation}\label{q3}
Avg_i=\frac{\sum_{j=1}^{H}\sum_{k=1}^{W} F_{i,j,k}}{H\times W}
\end{equation}
where $H$ and $W$ represent the rows and columns of $F_{i}$, respectively.

For client $p_i$, we further calculate the similarity deviation value $Dev$ by adding the normalized deviations $\triangle Dis$, $\triangle Cos$ and $\triangle Avg$,
as follows:
\begin{equation}\label{q4}
Dev_i= \frac{\triangle Dis_i}{\sum_{j=1}^{K}(\triangle Dis_{j})} + \frac{\triangle Cos_i}{\sum_{j=1}^{K}(\triangle Cos_{j})} +  \frac{\triangle Avg_i}{\sum_{j=1}^{K}(\triangle Avg_{j})}
\end{equation}

The reason why three metrics are used to calculate $Dev$ is to comprehensively consider various scenarios in that free-riders may exist,
and reduce the success rate of free-riders bypassing defenses.
Specifically,
Euclidean distance can be used to effectively identify free-riders,
but cannot work when the number of benign clients is close to free-riders
due to its symmetric nature.
Therefore, we leverage cosine similarity and average frequency to perform a better distinction.
These three metrics complement each other and work together.

The server sets the reputation threshold $\xi$ according to the similarity deviation value,
then separates benign clients and free-riders into $\left \{ P_n, P_r \right \}$.
Through experimental evaluation,
we find that the similarity deviation gap between benign clients and free-rider is large,
but the similarity deviation gap between free-riders is small.
Thus, free-riders can be identified by setting a certain range according to the maximum similarity deviation value.
We define $\xi\!=\!\max(Dev)\!-\!\epsilon$ in the experiment,
where $\epsilon$ is a hyperparameter.
We set $\epsilon=0.05$ by conducting a preliminary study based on a small dataset, 
and find that such a setting is effective in general.

\subsection{Personalized Model Aggregation}\label{ma}
Based on the client separation process,
the server can maintain two separated models in each round,
and aggregate the model updates uploaded by the two groups of clients respectively.
The sever leverage the two groups $\{P_{n},P_{r}\}$ to form two global models,
and then distribute them to the corresponding groups, respectively.
As a result, the global model trained by benign clients cannot be obtained by the free-riders.
The aggregation process is as follows:
\begin{equation}\label{j1}
\left \{ P_n \right \} : w^{t+1}_{g} =  w^t_g+ \frac{1}{\sum \left \{ P_n \right \} }  \sum_{i\in P_n}   \left ( w_i^{t+1}-w^t_g \right )
\end{equation}

\begin{equation}\label{j2}
\left \{ P_r \right \} : w^{t+1}_{g} =  w^t_g+ \frac{1}{\sum \left \{  P_r \right \}} \sum_{i\in P_r} \left ( w_i^{t+1}-w^t_g \right )
\end{equation}
where $w^{t+1}_g$ and $w^t_g$ are the global model in the ($t\!+\!1$)-th round and the $t$-th round,
$w_i^{t+1}$ is the local model updates uploaded by the client.
$\sum \left \{ P_n \right \}$ and $\sum \left \{ P_r \right \}$  represent the number of clients in their groups, respectively.

The server separates the clients in each round.
The evolving frequency of model weights collected by the server is continuously accumulated. Then, the difference in updated evolving frequency between the benign clients and the free-riders will be further enlarged.

The detailed implementation of WEF-Defense is shown in \textbf{Algorithm 1}.
It mainly includes three steps:
(1) WEF-Matrix information collection,
(2) clients separation,
and (3) personalized model aggregation.

\subsection{Algorithm Complexity}\label{Complexity}
We analyze the complexity of WEF-Defense in two parts, i.e., the information collection on the client and identification on the server.

On the client,
we select the weights of the penultimate layer in the model to initialize the WEF-Matrix, then use it to record the weight evolving frequency information. Therefore, the computational complexity can be defined as:
\begin{equation}
T_{client}  \sim  \mathcal{O}(1)  + \mathcal{O}(T')
\end{equation}
where $T'$ is the local training epochs.

On the server, we calculate $Dev$ and
perform model aggregation for clients in $\left \{ P_n,P_r \right \}$ respectively. Therefore, the time complexity is:
\begin{equation}
T_{server}  \sim  \mathcal{O}(K) + \mathcal{O}(K)
\end{equation}
where $K$ is the number of clients.

\begin{center}
\centering
\begin{tabular}{p{0.01\linewidth}p{0.9\linewidth}}
  \hline \hline
   & \textbf{Algorithm 1}: WEF-Defense. \\
  \hline
   & \textbf{Input}:
   dataset $\{D_i\}$ of each client,
   $i\in\{1,2,...,K\}$;
   the global epochs $T$ and the local epochs $T'$;
   initial global model $w_g^0$;
   total number of clients $K$;
   benign clients $P_n$ and free-riders $P_r$;
   hyperparameter $\epsilon=0.05$.\\
   & \textbf{Output}: the global models $\{P_n\}:w_g$ and $\{P_r\}:w_g$.\\
  \hline
  1. & Initialization: initialize WEF-Matrix based on Equ.~(\ref{equ:initialization}),
  local model $w_{i}^0 = w_g^0$.  \\
  2  &  \textbf{Role:}  Client $p_i$~~~~~~~~\emph{\#WEF-Matrix Information Collection} \\
  3. & ~~~~\textbf{If} {$i\in P_n $}\\
  4. & ~~~~~~~~$w_i^{t+1} \Leftarrow NormalUpdate(w_g^t)$\\
  5. & ~~~~~~~~\textbf{For} $t'<T'$ do\\
  6. & ~~~~~~~~~~~~Calculate $ F_i^{t'+1}$ according to Equ.~(\ref{th}) \\
  7. & ~~~~~~~~\textbf{End For} \\
  8. & ~~~~\textbf{Else If} {$i\in P_r $}\\
  9. & ~~~~~~~~$w_i^{t+1} \Leftarrow {FakeUpdates(w_g^t)}$\\
  10. & ~~~~~~~~\textbf{For} $t'<T'$ do \\
  11. & ~~~~~~~~~~~~Calculate $ F_i^{t'+1}$ according to Equ.~(\ref{th}) \\
  12. & ~~~~~~~~\textbf{End For} \\
  13. &~~~~ \textbf{End If}  \\
  14. &~~~~ {Local updates} ~$F_i^{t+1} ,w_i^{t+1}$~{upload to Server}\\
  15. & \textbf{Role:} Server~~~~~~~~\emph{\#Client Separation and Personalized Model Aggregation} \\
  16. &~~~~ \textbf{For} $F_i\in\{F_1,F_2,F_3,...,F_K\}$ \textbf{do}  \\
  17. &~~~~~~~~  {Calculate $ Dis_{i}$,$Cos_{i}$ and $Avg_{i}$ ~according~to~Equs.(\ref{q1}),(\ref{q2}) and (\ref{q3}) }\\
  18. &~~~~ \textbf{End For} \\
  19. &~~~~ \textbf{For} $i < K$ \textbf{do}  \\
  20. & ~~~~~~~~  Calculate~$Dev_{i}$ ~according to Equ.~(\ref{q4})   \\
  21. &~~~~ \textbf{End For} \\
  22. &~~~~ Separate clients to groups $\left \{ P_n, P_r \right \}$ according to the reputation threshold. \\
  23. &~~~~~Calculate $\{P_n\}:w_g^{t+1}$ and $\{P_r\}:w_g^{t+1}$ according to Equs.~(\ref{j1}) and (\ref{j2})  \\
  24. & \textbf{Return}: the global models $\{P_n\}:w_g$ and $\{P_r\}:w_g$ \\
  \hline \hline
\end{tabular}
\end{center}

\section{Experiments Setting }\label{setting}

\textbf{Platform}:
i7-7700K 4.20GHzx8 (CPU), TITAN Xp 12GiB x2 (GPU), 16GBx4 memory (DDR4), Ubuntu 16.04 (OS), Python 3.6, pytorch1.8.2~\cite{paszke2017automatic}.

\textbf{Datasets}: We evaluate WEF-Defense on five datasets,
i.e., MNIST, CIFAR-10, GTSRB, BANK and ADULT.
MNIST~\cite{lecun1998gradient} dataset contains 70,000 real-world handwritten images with digits ranging from 0 to 9.
CIFAR-10~\cite{krizhevsky2009learning} dataset contains 60,000 color images in 10 classes with a size of 32 x 32 and 6,000 images per class.
GTSRB~\cite{sermanet2011traffic} dataset contains 51,839 real-world German colored traffic signs in 43 categories.
In ADULT~\cite{kohavi1996scaling} dataset, there are 48,843 records in total.
We manually balance the ADULT dataset to have
11,687 records over 50K and 11,687 records under 50K, resulting in total of 23,374 records.
BANK~\cite{moro2011using} dataset is related to the direct marketing campaign of a Portuguese banking institution and has data on whether 45,211 customers subscribe to fixed deposits, each with 16 attributes.
For each dataset, we conduct an 80-20 train-test split.
The detailed information of datasets is shown in Table~\ref{table1}.

\begin{table}[htb]
\centering
\caption{Dataset and model parameter settings.}\label{table1}
\resizebox{0.90\linewidth}{!}{
\begin{tabular}{ccccccccc}
\hline\hline
\textbf{Datasets}       &\textbf{Samples} &\textbf{Dimensions}  &\textbf{Classes}   & \textbf{Models}    &\textbf{LearningRate}      &\textbf{Momentum}    &\textbf{Epoches} &\textbf{BachSize}         \\ \hline
MNIST      & 70,000  &28$\times$28   & 10        & LeNet  & 0.005    &0.0001   &50      & 32           \\
CIFAR-10   & 60,000  &32$\times$32   & 10       & VGG16  & 0.01  &0.9     &80      & 32         \\
GTSRB      &51,839  &32$\times$32    &43          &ResNet18   &0.001   &0.001        &80      & 32         \\
ADULT     &23,374   &14   &2           &MLP  &0.0001   &0.0001        &50      & 32  \\
BANK       &45,211  &16   &2            &MLP  &0.02   &0.5        &80      & 32
  \\
\hline\hline
\end{tabular}}
\end{table}

\textbf{Data Distribution}: Two typical data distribution scenarios are considered in our experiments.
Independent and identically distribute (IID) data~\cite{journals/sac/RachedHRT21}:
each client contains the same amount of data,
and contains complete categories.
Non-independent and identically distribute (Non-IID) data~\cite{journals/ijon/ZhuXLJ21}:
in real-world scenarios,
the data among clients is heterogeneous,
we consider using Dirichlet distribution~\cite{journals/tnn/ZamzamiB22,conf/ssp/RademacherD21,ZoghbiVM16} to divide the training data among clients.
Specifically, we sample $Dir(\beta)$
and divide the dataset according to the distribution of concentration parameter $\beta$, assigned to each client.
More specific, $Dir(\beta)$ is the Dirichlet distribution with $\beta$.
With the above partitioning strategy,
each client can have relatively few data samples in certain classes.
Consider using $\beta$=0.5 in the experiment to explore the problem of heterogeneity.

\textbf{Number of clients}: In all experimental scenarios,
we mainly evaluate the effect of different ratios of free-riders on our method.
Thus the total number of clients is 10, and the free-rider attacks are discussed for 10\%, 30\%, 50\% and 90\% of the total clients.

\textbf{Models}:
Different classifiers are used for various datasets.
For MNIST, LeNet~\cite{el2016cnn} is used for classification.
For more complex image datasets,
CIFAR-10 and GTSRB,
VGG16~\cite{simonyan2014very} and ResNet18~\cite{he2016deep} are adopted, respectively.
For structured datasets,
ADULT and ADULT BANK,
MLP~\cite{karlik2011performance} is applied.
Refer to Table \ref{table1} for specific parameter setting.
All evaluation results are the average of 3 runs of the same setting.

\textbf{Hyper-Parameters}:
For all experiments, we set the hyperparameter $\epsilon=0.05$.


\textbf{Attack Methods}: Three existing free-rider attack methods are applied to evaluate the detection performance,
including random weight attack~\cite{lin2019free},
stochastic perturbations attack~\cite{lin2019free} and
delta weight attack~\cite{lin2019free}.
Among them,
the weight generation range $R$ of random weight attack uses $10^{-3}$.
In the adaptive attack scenario,
we design a new free-rider attack to evaluate the defense performance.

\textbf{Baselines}: Two defense approaches are used for comparison,
including CFFL~\cite{lyu2020collaborative} based on the validation dataset,
and RFFL~\cite{xu2021reputation} based on cosine similarity between local gradients and aggregated global gradients.
The undefended FedAvg aggregation algorithm~\cite{mcmahan2017communication} as a benchmark.

\textbf{Evaluation Metrics}:
We evaluate the performance of the detection methods by evaluating the highest mean accuracy (HMA) of the model that can be stolen by free-riders.
The lower the HMA is, the better the detection is.

\section{Evaluation and Analysis}\label{eval}
In this section, we evaluate the performance of WEF-Defense by answering the following five research questions (RQs):

\begin{itemize}
\item \textbf{RQ1}: Dose WEF-Defense achieves the SOTA defense performance compared with baselines when defending against various free-rider attacks?

\item \textbf{RQ2}: Does WEF-Defense still achieve the best performance when the proportion of free-riders is higher?

\item \textbf{RQ3}: Will WEF-Defense affect the main task performance? What is its communication overhead?

\item \textbf{RQ4}: How to interpret the defense of WEF-Defense through visualizations?

\item \textbf{RQ5}: Can WEF-Defense defend against adaptive attack? How sensitive is the hyperparameter?
\end{itemize}

\subsection{RQ1: Defense Effectiveness of WEF-Defense}\label{sec4}
In this section, 
we verify the defense effect of WEF-Defense compared with baselines on different datasets for different models.

\begin{table}[htb]\label{tab2}
\caption{The HMA (\%) comparison between WEF-Defense and baselines under the IID setting,
where ``10\%'' and ``30\%'' represent different free-rider ratios. 
}
\centering
\resizebox{0.8\textwidth}{!}{%
\begin{tabular}{cccccccccc}
\hline\hline
\multicolumn{1}{l}{\multirow{2}{*}{\textbf{IID Datasets}}} & \multicolumn{1}{l}{\multirow{2}{*}{\textbf{Attacks}}} & \multicolumn{2}{c}{\textbf{FedAvg}}                         & \multicolumn{2}{c}{\textbf{CFFL}  }                                     & \multicolumn{2}{c}{\textbf{RFFL}}                                       & \multicolumn{2}{c}{\textbf{WEF-Defense}}                             \\
\multicolumn{1}{l}{}                                 & \multicolumn{1}{l}{}                        & \multicolumn{1}{c}{\textbf{10\%}} & \multicolumn{1}{c|}{\textbf{30\%}}  & \multicolumn{1}{c}{\textbf{10\%}} & \multicolumn{1}{c|}{\textbf{30\%}}           & \multicolumn{1}{c}{\textbf{10\%}} & \multicolumn{1}{c|}{\textbf{30\%}}           & \multicolumn{1}{c}{\textbf{10\%}} & \multicolumn{1}{c}{\textbf{30\%}} \\ \hline
\multicolumn{1}{c|}{\multirow{3}{*}{MNIST}}      & \multicolumn{1}{c|}{RWA}                     & 10.75                    & \multicolumn{1}{r|}{10.81} & 10.75                    & \multicolumn{1}{r|}{10.75}          & 10.75                    & \multicolumn{1}{r|}{10.85} & \textbf{10.13}           & \textbf{9.91}            \\
\multicolumn{1}{c|}{}                            & \multicolumn{1}{c|}{SPA}                     & 98.26                    & \multicolumn{1}{r|}{97.92} & 81.35                    & \multicolumn{1}{r|}{71.83}          & 55.01                    & \multicolumn{1}{r|}{68.41} & \textbf{9.43}            & \textbf{10.71}           \\
\multicolumn{1}{c|}{}                            & \multicolumn{1}{c|}{DWA}                     & 98.05                    & \multicolumn{1}{r|}{98.36} & 77.52                    & \multicolumn{1}{r|}{87.81}          & 87.74                    & \multicolumn{1}{r|}{79.21} & \textbf{9.52}            & \textbf{22.21}           \\ \hline
\multicolumn{1}{c|}{\multirow{3}{*}{CIFAR-10}}   & \multicolumn{1}{c|}{RWA}                     & 44.78                    & \multicolumn{1}{r|}{10.55} & 10.31           & \multicolumn{1}{r|}{11.30}          & 10.65                    & \multicolumn{1}{r|}{10.55} & \textbf{10.12}           & \textbf{10.41}           \\
\multicolumn{1}{c|}{}                            & \multicolumn{1}{c|}{SPA}                     & 84.55                    & \multicolumn{1}{r|}{84.12} & 68.25                    & \multicolumn{1}{r|}{75.53}          & 10.05                    & \multicolumn{1}{r|}{10.85} & \textbf{9.11}            & \textbf{10.66}           \\
\multicolumn{1}{c|}{}                            & \multicolumn{1}{c|}{DWA}                     & 86.45                    & \multicolumn{1}{r|}{84.92} & 56.43                    & \multicolumn{1}{r|}{59.45}          & 20.59                    & \multicolumn{1}{r|}{20.72} & \textbf{9.42}            & \textbf{19.74}           \\ \hline
\multicolumn{1}{c|}{\multirow{3}{*}{GTSRB}}      & \multicolumn{1}{c|}{RWA}                     & 72.40                    & \multicolumn{1}{r|}{6.13}  & 4.86                     & \multicolumn{1}{r|}{4.98}           & 5.06                     & \multicolumn{1}{r|}{5.06}  & \textbf{3.56}            & \textbf{4.86}            \\
\multicolumn{1}{c|}{}                            & \multicolumn{1}{c|}{SPA}                     & 94.57                    & \multicolumn{1}{r|}{94.12} & \textbf{4.70}            & \multicolumn{1}{r|}{4.71}           & 4.71                     & \multicolumn{1}{r|}{4.71}  & 4.86                     & \textbf{3.52}            \\
\multicolumn{1}{c|}{}                            & \multicolumn{1}{c|}{DWA}                     & 94.65                    & \multicolumn{1}{r|}{94.18} & 39.66                    & \multicolumn{1}{r|}{39.19}          & 39.77                    & \multicolumn{1}{r|}{38.36} & \textbf{35.27}           & \textbf{36.38}           \\ \hline
\multicolumn{1}{c|}{\multirow{3}{*}{ADULT}}      & \multicolumn{1}{c|}{RWA}                     & 50.00                    & \multicolumn{1}{r|}{50.00} & 76.83                    & \multicolumn{1}{r|}{\textbf{49.44}} & 74.93                    & \multicolumn{1}{r|}{57.85} & \textbf{50.00}           & 49.96                    \\
\multicolumn{1}{c|}{}                            & \multicolumn{1}{c|}{SPA}                     & 78.88                    & \multicolumn{1}{r|}{78.91} & 78.95                    & \multicolumn{1}{r|}{77.24}          & 79.21                    & \multicolumn{1}{r|}{52.66} & \textbf{35.22}           & \textbf{45.34}           \\
\multicolumn{1}{c|}{}                            & \multicolumn{1}{c|}{DWA}                     & 78.92                    & \multicolumn{1}{r|}{78.91} & 78.91                    & \multicolumn{1}{r|}{78.96}          & 73.19                    & \multicolumn{1}{r|}{73.79} & \textbf{61.26}           & \textbf{56.23}           \\ \hline
\multicolumn{1}{c|}{\multirow{3}{*}{BANK}}       & \multicolumn{1}{c|}{RWA}                     & 84.56                    & \multicolumn{1}{r|}{71.25} & 79.95                    & \multicolumn{1}{r|}{74.65}          & 50.24                    & \multicolumn{1}{r|}{50.10} & \textbf{50.01}           & \textbf{50.00}           \\
\multicolumn{1}{c|}{}                            & \multicolumn{1}{c|}{SPA}                     & 84.66                    & \multicolumn{1}{r|}{83.85} & 80.55                    & \multicolumn{1}{r|}{79.14}          & 65.64                    & \multicolumn{1}{r|}{67.40} & \textbf{49.95}           & \textbf{50.00}           \\
\multicolumn{1}{c|}{}                            & \multicolumn{1}{c|}{DWA}                     & 84.63                    & \multicolumn{1}{r|}{82.88} & 75.94                    & \multicolumn{1}{r|}{74.65}          & 71.95                    & \multicolumn{1}{r|}{72.45} & \textbf{63.52}           & \textbf{70.00}           \\ 
\hline\hline
\end{tabular}
}\label{tab2}
\end{table}

\textbf{Implementation Details.}
(1) Five datasets are tested in IID data and Non-IID data settings. The Non-IID data adopts the Dirichlet distribution to explore the problem of heterogeneity, where the distribution coefficient defaults to 0.5.
(2) In general, the number of free-riders is less than that of benign clients. Consequently, in 10 clients, two scenarios with the free-rider ratio of 10\% and 30\% are set up, in which the camouflage method of free-rider adopts random weight attack (RWA), stochastic perturbations attack (SPA) and delta weight attack (DWA).
(3) We adopt three baselines to perform the comparison, 
i.e., undefended FedAvg aggregation~\cite{mcmahan2017communication},
RFFL~\cite{xu2021reputation} and 
CFFL~\cite{lyu2020collaborative}. 
We use the HMA obtained by free-rider as the evaluation metric. 
The results of IID data and Non-IID data are shown in Tables~\ref{tab2} and~\ref{tab3noniid}, respectively.

\textbf{Results and Analysis.}
The results in Tables~\ref{tab2} and~\ref{tab3noniid} demonstrate that 
the overall defense performance of WEF-Defense is the best compared with CFFL and RFFL in most cases.
Particularly, 
the defense effect of WEF-Defense remains stable when dealing with the heterogeneity of Non-IID data. 
For all image datasets (i.e., MNIST, CIFAR-10 and GTSRB) in both tables, 
WEF-Defense makes the overall HMA obtained by free-riders not exceed 36.38\%, 
where the lowest HMA is only 2.29\%. 
For all structured datasets (i.e., ADULT and BANK) in both tables, 
WEF-Defense realizes the overall HMA obtained by free-riders not exceed 70\%, 
among which the lowest HMA is only 35.22\%. 
It is worth noting that 
since there are only two classes in both ADULT and BANK datasets, the HMA should reach about 50\% during the initial training, 
i.e., random guessing. 
Therefore, we can conclude that 
WEF-Defense performs the SOTA defense compared with baselines.


\begin{table}[htb]
\caption{
The HMA (\%) comparison between WEF-Defense and baselines under the Non-IID setting,
where ``10\%'' and ``30\%'' represent different free-rider ratios. 
}\label{tab3noniid}
\centering
\resizebox{0.8\textwidth}{!}{%
\begin{tabular}{cccccccccc}
\hline\hline
\multicolumn{1}{l}{\multirow{2}{*}{\textbf{Non-IID Datasets}}} & \multicolumn{1}{l}{\multirow{2}{*}{\textbf{Attacks}}} & \multicolumn{2}{c}{\textbf{FedAvg}}                         & \multicolumn{2}{c}{\textbf{CFFL}  }                                     & \multicolumn{2}{c}{\textbf{RFFL}}                                       & \multicolumn{2}{c}{\textbf{WEF-Defense}}                             \\
\multicolumn{1}{l}{}                                 & \multicolumn{1}{l}{}                        & \multicolumn{1}{c}{\textbf{10\%}} & \multicolumn{1}{c|}{\textbf{30\%}}  & \multicolumn{1}{c}{\textbf{10\%}} & \multicolumn{1}{c|}{\textbf{30\%}}           & \multicolumn{1}{c}{\textbf{10\%}} & \multicolumn{1}{c|}{\textbf{30\%}}           & \multicolumn{1}{c}{\textbf{10\%}} & \multicolumn{1}{c}{\textbf{30\%}} \\ \hline
\multicolumn{1}{c|}{\multirow{3}{*}{MNIST}}          & \multicolumn{1}{c|}{RWA}                     & 78.41                    & \multicolumn{1}{r|}{76.55} & 25.03                    & \multicolumn{1}{r|}{14.49}         & 10.73                    & \multicolumn{1}{r|}{10.81} & \textbf{10.02}           & \textbf{9.99}            \\
\multicolumn{1}{c|}{}                                & \multicolumn{1}{c|}{SPA}                     & 95.22                    & \multicolumn{1}{r|}{97.01} & 68.41                    & \multicolumn{1}{r|}{59.75}         & 17.92                    & \multicolumn{1}{r|}{26.70} & \textbf{9.72}            & \textbf{10.76}           \\
\multicolumn{1}{c|}{}                                & \multicolumn{1}{c|}{DWA}                     & 95.10                    & \multicolumn{1}{r|}{98.54} & 40.42                    & \multicolumn{1}{r|}{64.92}         & 71.31                    & \multicolumn{1}{r|}{62.64} & \textbf{20.22}           & \textbf{13.91}           \\ \hline
\multicolumn{1}{c|}{\multirow{3}{*}{CIFAR-10}}       & \multicolumn{1}{c|}{RWA}                     & 53.91                    & \multicolumn{1}{r|}{9.06}  & 9.52                     & \multicolumn{1}{r|}{10.04}         & 9.50                     & \multicolumn{1}{r|}{11.32} & \textbf{9.05}            & \textbf{9.91}            \\
\multicolumn{1}{c|}{}                                & \multicolumn{1}{c|}{SPA}                     & 84.83                    & \multicolumn{1}{r|}{85,42} & 28.54                    & \multicolumn{1}{r|}{16.72}         & 9.02                     & \multicolumn{1}{r|}{11.65} & \textbf{8.92}            & \textbf{9.52}            \\
\multicolumn{1}{c|}{}                                & \multicolumn{1}{c|}{DWA}                     & 84.44                    & \multicolumn{1}{r|}{84.98} & 19.12                    & \multicolumn{1}{r|}{33.62}         & 17.52                    & \multicolumn{1}{r|}{23.13} & \textbf{9.45}            & \textbf{20.00}           \\ \hline
\multicolumn{1}{c|}{\multirow{3}{*}{GTSRB}}          & \multicolumn{1}{c|}{RWA}                     & 6.13                     & \multicolumn{1}{r|}{5.82}  & 6.13                     & \multicolumn{1}{r|}{5.85}          & 5.90                     & \multicolumn{1}{r|}{4.52}  & \textbf{4.71}            & \textbf{4.86}            \\
\multicolumn{1}{c|}{}                                & \multicolumn{1}{c|}{SPA}                     & 94.00                    & \multicolumn{1}{r|}{92.63} & 4.71                     & \multicolumn{1}{r|}{\textbf{4.71}} & 4.72                     & \multicolumn{1}{r|}{4.75}  & \textbf{2.29}            & 4.82                     \\
\multicolumn{1}{c|}{}                                & \multicolumn{1}{c|}{DWA}                     & 94.22                    & \multicolumn{1}{r|}{94.00} & \textbf{23.35}           & \multicolumn{1}{r|}{24.86}         & 28.43                    & \multicolumn{1}{r|}{28.73} & 30.04                    & \textbf{19.11}           \\ \hline
\multicolumn{1}{c|}{\multirow{3}{*}{ADULT}}          & \multicolumn{1}{c|}{RWA}                     & 50.03                    & \multicolumn{1}{r|}{50.04} & 50.20                    & \multicolumn{1}{r|}{50.10}         & 50.01                    & \multicolumn{1}{r|}{49.94} & \textbf{49.62}           & \textbf{49.92}           \\
\multicolumn{1}{c|}{}                                & \multicolumn{1}{c|}{SPA}                     & 76.22                    & \multicolumn{1}{r|}{76.02} & 59.92                    & \multicolumn{1}{r|}{55.42}         & 54.91                    & \multicolumn{1}{r|}{60.00} & \textbf{49.62}           & \textbf{46.42}           \\
\multicolumn{1}{c|}{}                                & \multicolumn{1}{c|}{DWA}                     & 77.00                    & \multicolumn{1}{r|}{78.36} & 63.62                    & \multicolumn{1}{r|}{57.42}         & \textbf{49.92}           & \multicolumn{1}{r|}{50.05} & 51.82                    & \textbf{49.92}           \\ \hline
\multicolumn{1}{c|}{\multirow{3}{*}{BANK}}           & \multicolumn{1}{c|}{RWA}                     & 71.21                    & \multicolumn{1}{r|}{50.02} & 50.23                    & \multicolumn{1}{r|}{50.21}         & 50.54                    & \multicolumn{1}{r|}{50.10} & \textbf{50.00}           & \textbf{50.00}           \\
\multicolumn{1}{c|}{}                                & \multicolumn{1}{c|}{SPA}                     & 76.95                    & \multicolumn{1}{r|}{71.75} & 70.45                    & \multicolumn{1}{r|}{50.99}         & 50.13                    & \multicolumn{1}{r|}{50.92} & \textbf{50.00}           & \textbf{50.00}           \\
\multicolumn{1}{c|}{}                                & \multicolumn{1}{c|}{DWA}                     & 80.63                    & \multicolumn{1}{r|}{80.26} & 50.52                    & \multicolumn{1}{r|}{71.05}         & 57.09                    & \multicolumn{1}{r|}{71.05} & \textbf{50.00}           & \textbf{50.00}           \\ \hline\hline
\end{tabular}}
\end{table}

Besides,
WEF-Defense shows more stable performance in defending against various free-rider attacks in different scenarios.
For instance,
in Table~\ref{tab2} with the IID setting,
the standard deviation of HMA for WEF-Defense on image datasets is around 9.36, 
while that for CFFL and RFFL reaches 31.19 and 26.72, respectively.
In Table~\ref{tab3noniid} with the Non-IID setting, 
the standard deviation of HMA for WEF-Defense is around 6.69, 
while that for CFFL and RFFL reaches 20.04 and 18.43, respectively.
The average standard deviation of WEF-Defense is about 1/3 that of baselines, 
which further demonstrates its stability.


\begin{center}
\fcolorbox{black}{white!20}{\parbox{0.97\linewidth} {
\emph{\textbf{Answer to RQ1:}}
WEF-Defense shows the SOTA performance compared with baselines and prevents various free-rider attacks,
whether 10\% or 30\% of clients are free-riders.
Under the IID and Non-IID settings, on average,
1)~its defense effect is 1.68 and 1.33 times that of baselines, respectively; and 
2)~its defense stability is 3.09 and 2.87 times that of baselines, respectively.
    }
}
\end{center}

\subsection{RQ2: Defense Effect at Higher Free-Rider Ratios}\label{sec4}
Under the traditional FL framework, 
more than half of the total clients of free-riders do not have much impact on the global model's accuracy. 
For instance,
in Table~\ref{tab3}, 
free-riders with DWA realize over 80\% HMA on average 
when the number of free-riders reaches 90\% of all clients
in the undefended FedAvg aggregation framework.
Therefore, we consider whether a high proportion of free-riders affects defense effectiveness.

\begin{table}[htb]\label{tab3}
\centering
\caption{
The HMA (\%) comparison between WEF-Defense and baselines under the IID setting,
where ``50\%'' and ``90\%'' represent different free-rider ratios. 
}
\resizebox{0.8\textwidth}{!}{%
\begin{tabular}{cccccccccc}
\hline\hline
\multicolumn{1}{l}{\multirow{2}{*}{\textbf{IID Datasets}}} & \multicolumn{1}{l}{\multirow{2}{*}{\textbf{Attacks}}} & \multicolumn{2}{c}{\textbf{FedAvg}}                         & \multicolumn{2}{c}{\textbf{CFFL}  }                                     & \multicolumn{2}{c}{\textbf{RFFL}}                                       & \multicolumn{2}{c}{\textbf{WEF-Defense}}                             \\
\multicolumn{1}{l}{}                                 & \multicolumn{1}{l}{}                        & \multicolumn{1}{c}{\textbf{50\%}} & \multicolumn{1}{c|}{\textbf{90\%}}  & \multicolumn{1}{c}{\textbf{50\%}} & \multicolumn{1}{c|}{\textbf{90\%}}           & \multicolumn{1}{c}{\textbf{50\%}} & \multicolumn{1}{c|}{\textbf{90\%}}           & \multicolumn{1}{c}{\textbf{50\%}} & \multicolumn{1}{c}{\textbf{90\%}} \\ \hline
\multicolumn{1}{c|}{\multirow{3}{*}{MNIST}}      & \multicolumn{1}{c|}{RWA}                     & 10.82                    & \multicolumn{1}{r|}{10.23} & 10.65                    & \multicolumn{1}{r|}{\textbf{9.34}}  & 14.39                    & \multicolumn{1}{r|}{10.65} & \textbf{8.71}            & 10.73                    \\
\multicolumn{1}{c|}{}                            & \multicolumn{1}{c|}{SPA}                     & 97.44                    & \multicolumn{1}{r|}{89.24} & 66.34                    & \multicolumn{1}{r|}{64.34}          & 76.52                    & \multicolumn{1}{r|}{22.84} & \textbf{10.72}           & \textbf{10.72}           \\
\multicolumn{1}{c|}{}                            & \multicolumn{1}{c|}{DWA}                     & 98.25                    & \multicolumn{1}{r|}{95.73} & 75.16                    & \multicolumn{1}{r|}{38.45}          & 87.74                    & \multicolumn{1}{r|}{79.21} & \textbf{19.82}           & \textbf{27.22}           \\ \hline
\multicolumn{1}{c|}{\multirow{3}{*}{CIFAR-10}}   & \multicolumn{1}{c|}{RWA}                     & 11.35                    & \multicolumn{1}{r|}{10.72} & 11.31                    & \multicolumn{1}{r|}{9.90}           & 11.30                    & \multicolumn{1}{r|}{11.30} & \textbf{10.21}           & \textbf{9.30}            \\
\multicolumn{1}{c|}{}                            & \multicolumn{1}{c|}{SPA}                     & 80.00                    & \multicolumn{1}{r|}{45.53} & 59.49                    & \multicolumn{1}{r|}{\textbf{10.05}} & 10.85                    & \multicolumn{1}{r|}{11.75} & \textbf{10.60}           & 11.38                    \\
\multicolumn{1}{c|}{}                            & \multicolumn{1}{c|}{DWA}                     & 82.55                    & \multicolumn{1}{r|}{65.96} & 74.67                    & \multicolumn{1}{r|}{23.39}          & 21.63                    & \multicolumn{1}{r|}{22.45} & \textbf{21.30}           & \textbf{21.12}           \\ \hline
\multicolumn{1}{c|}{\multirow{3}{*}{GTSRB}}      & \multicolumn{1}{c|}{RWA}                     & 6.13                     & \multicolumn{1}{r|}{5.06}  & 6.13                     & \multicolumn{1}{r|}{6.13}           & 5.85                     & \multicolumn{1}{r|}{6.13}  & \textbf{4.94}            & \textbf{4.86}            \\
\multicolumn{1}{c|}{}                            & \multicolumn{1}{c|}{SPA}                     & 93.74                    & \multicolumn{1}{r|}{84.36} & \textbf{4.71}            & \multicolumn{1}{r|}{4.87}           & 4.73                     & \multicolumn{1}{r|}{4.91}  & 4.86                     & \textbf{4.86}            \\
\multicolumn{1}{c|}{}                            & \multicolumn{1}{c|}{DWA}                     & 93.34                    & \multicolumn{1}{r|}{84.52} & {36.74}           & \multicolumn{1}{r|}{\textbf{29.17}} & 39.58                    & \multicolumn{1}{r|}{35.41} & \textbf{36.10}                    & 36.50                    \\ \hline
\multicolumn{1}{c|}{\multirow{3}{*}{ADULT}}      & \multicolumn{1}{c|}{RWA}                     & 50.24                    & \multicolumn{1}{r|}{50.00} & 50.15                    & \multicolumn{1}{r|}{50.10}          & 53.61                    & \multicolumn{1}{r|}{51.20} & \textbf{50.00}           & \textbf{49.91}           \\
\multicolumn{1}{c|}{}                            & \multicolumn{1}{c|}{SPA}                     & 78.66                    & \multicolumn{1}{r|}{72.66} & 79.14                    & \multicolumn{1}{r|}{63.12}          & 68.11                    & \multicolumn{1}{r|}{54.12} & \textbf{45.52}           & \textbf{47.42}           \\
\multicolumn{1}{c|}{}                            & \multicolumn{1}{c|}{DWA}                     & 79.11                    & \multicolumn{1}{r|}{79.15} & 78.99                    & \multicolumn{1}{r|}{78.94}          & 73.43                    & \multicolumn{1}{r|}{72.92} & \textbf{56.13}           & \textbf{56.32}           \\ \hline
\multicolumn{1}{c|}{\multirow{3}{*}{BANK}}       & \multicolumn{1}{c|}{RWA}                     & 51.03                    & \multicolumn{1}{r|}{51.40} & 50.15                    & \multicolumn{1}{r|}{50.60}          & 50.30                    & \multicolumn{1}{r|}{50.30} & \textbf{50.14}           & \textbf{50.00}           \\
\multicolumn{1}{c|}{}                            & \multicolumn{1}{c|}{SPA}                     & 83.54                    & \multicolumn{1}{r|}{80.85} & 76.85                    & \multicolumn{1}{r|}{66.85}          & 70.65                    & \multicolumn{1}{r|}{69.39} & \textbf{50.00}           & \textbf{50.00}           \\
\multicolumn{1}{c|}{}                            & \multicolumn{1}{c|}{DWA}                     & 83.33                    & \multicolumn{1}{r|}{83.45} & 72.39                    & \multicolumn{1}{r|}{69.25}          & 72.45                    & \multicolumn{1}{r|}{72.35} & \textbf{68.89}           & \textbf{68.53}           \\ \hline
\hline
\end{tabular}
}\label{tab3}
\end{table}

\textbf{Implementation Details.} (1) The IID and Non-IID are adopted for the five datasets, respectively. 
The Non-IID data adopts the Dirichlet distribution to explore the problem of heterogeneity, where the distribution coefficient defaults to 0.5. 
(2) We set the free-rider ratio to 50\% and 90\% in 10 clients. 
It helps to discover how WEF-Defense performs when the number of free-riders is equal to or much more than that of benign clients.
Tables \ref{tab3} and \ref{tab5noniid} show the results.

\begin{table}[htb]
\centering
\caption{
The HMA (\%) comparison between WEF-Defense and baselines under the Non-IID setting,
where ``50\%'' and ``90\%'' represent different free-rider ratios. 
}\label{tab5noniid}
\resizebox{0.8\textwidth}{!}{%
\begin{tabular}{cccccccccc}
\hline\hline
\multicolumn{1}{l}{\multirow{2}{*}{\textbf{Non-IID Datasets}}} & \multicolumn{1}{l}{\multirow{2}{*}{\textbf{Attacks}}} & \multicolumn{2}{c}{\textbf{FedAvg}}                         & \multicolumn{2}{c}{\textbf{CFFL}  }                                     & \multicolumn{2}{c}{\textbf{RFFL}}                                       & \multicolumn{2}{c}{\textbf{WEF-Defense}}                             \\
\multicolumn{1}{l}{}                                 & \multicolumn{1}{l}{}                        & \multicolumn{1}{c}{\textbf{50\%}} & \multicolumn{1}{c|}{\textbf{90\%}}  & \multicolumn{1}{c}{\textbf{50\%}} & \multicolumn{1}{c|}{\textbf{90\%}}           & \multicolumn{1}{c}{\textbf{50\%}} & \multicolumn{1}{c|}{\textbf{90\%}}           & \multicolumn{1}{c}{\textbf{50\%}} & \multicolumn{1}{c}{\textbf{90\%}} \\ \hline
\multicolumn{1}{c|}{\multirow{3}{*}{MNIST}}          & \multicolumn{1}{c|}{RWA}                     & 58.13                    & \multicolumn{1}{r|}{9.48}  & 26.23                    & \multicolumn{1}{r|}{10.84}          & 18.81                    & \multicolumn{1}{r|}{10.81}          & \textbf{10.81}           & \textbf{9.98}            \\
\multicolumn{1}{c|}{}                                & \multicolumn{1}{c|}{SPA}                     & 96.86                    & \multicolumn{1}{r|}{98.94} & 58.42                    & \multicolumn{1}{r|}{16.87}          & 62.92                    & \multicolumn{1}{r|}{33.89}          & \textbf{10.72}           & \textbf{10.72}           \\
\multicolumn{1}{c|}{}                                & \multicolumn{1}{c|}{DWA}                     & 97.75                    & \multicolumn{1}{r|}{99.21} & 55.81                    & \multicolumn{1}{r|}{82.89}          & 61.41                    & \multicolumn{1}{r|}{39.92}          & \textbf{21.43}           & \textbf{23.45}           \\ \hline
\multicolumn{1}{c|}{\multirow{3}{*}{CIFAR-10}}       & \multicolumn{1}{c|}{RWA}                     & 10.00                    & \multicolumn{1}{r|}{9.93}  & 10.00                    & \multicolumn{1}{r|}{\textbf{10.71}} & 11.39                    & \multicolumn{1}{r|}{11.35}          & \textbf{9.05}            & 11.31                    \\
\multicolumn{1}{c|}{}                                & \multicolumn{1}{c|}{SPA}                     & 85.87                    & \multicolumn{1}{r|}{79.53} & 10.00                    & \multicolumn{1}{r|}{10.32}          & \textbf{10.19}           & \multicolumn{1}{r|}{\textbf{10.23}} & 11.32                    & 11.33                    \\
\multicolumn{1}{c|}{}                                & \multicolumn{1}{c|}{DWA}                     & 85.73                    & \multicolumn{1}{r|}{85.51} & 29.25                    & \multicolumn{1}{r|}{40.52}          & 24.63                    & \multicolumn{1}{r|}{19.12}          & \textbf{19.18}           & \textbf{16.84}           \\ \hline
\multicolumn{1}{c|}{\multirow{3}{*}{GTSRB}}          & \multicolumn{1}{c|}{RWA}                     & 5.97                     & \multicolumn{1}{r|}{6.13}  & 6.13                     & \multicolumn{1}{r|}{5.85}           & 5.80                     & \multicolumn{1}{r|}{6.14}           & \textbf{5.46}            & \textbf{4.86}            \\
\multicolumn{1}{c|}{}                                & \multicolumn{1}{c|}{SPA}                     & 89.52                    & \multicolumn{1}{r|}{57.63} & {4.91}            & \multicolumn{1}{r|}{{4.92}}  & 5.70                     & \multicolumn{1}{r|}{4.83}           & \textbf{4.81 }                    & \textbf{4.80}                     \\
\multicolumn{1}{c|}{}                                & \multicolumn{1}{c|}{DWA}                     & 91.17                    & \multicolumn{1}{r|}{68.82} & \textbf{27.67}           & \multicolumn{1}{r|}{23.47}          & 32.23                    & \multicolumn{1}{r|}{28.94}          & 31.23                    & \textbf{22.73}           \\ \hline
\multicolumn{1}{c|}{\multirow{3}{*}{ADULT}}          & \multicolumn{1}{c|}{RWA}                     & 50.20                    & \multicolumn{1}{r|}{50.00} & 50.42                    & \multicolumn{1}{r|}{ 50.00}          & 52.45                   & \multicolumn{1}{r|}{51.18}          & \textbf{50.00}           & \textbf{49.72}                  \\
\multicolumn{1}{c|}{}                                & \multicolumn{1}{c|}{SPA}                     & 76.34                    & \multicolumn{1}{r|}{69.42} & 78.34                    & \multicolumn{1}{r|}{71.12}          & 61.04                    & \multicolumn{1}{r|}{55.23}          & \textbf{54.09}           & \textbf{49.79}           \\
\multicolumn{1}{c|}{}                                & \multicolumn{1}{c|}{DWA}                     & 79.05                    & \multicolumn{1}{r|}{79.00} & 73.34                    & \multicolumn{1}{r|}{79.42}          & 57.52           & \multicolumn{1}{r|}{52.16}          & \textbf{53.00}                    & \textbf{51.62}           \\ \hline
\multicolumn{1}{c|}{\multirow{3}{*}{BANK}}           & \multicolumn{1}{c|}{RWA}                     & 50.30                    & \multicolumn{1}{r|}{50.04} & 50.40                   & \multicolumn{1}{r|}{50.26}          & 50.19                    & \multicolumn{1}{r|}{50.20}          & \textbf{50.00}           & \textbf{50.00}           \\
\multicolumn{1}{c|}{}                                & \multicolumn{1}{c|}{SPA}                     & 83.33                    & \multicolumn{1}{r|}{50.02} & 55.15                    & \multicolumn{1}{r|}{53.84}          & 52.42                    & \multicolumn{1}{r|}{57.49}          & \textbf{50.00}           & \textbf{50.00}           \\
\multicolumn{1}{c|}{}                                & \multicolumn{1}{c|}{DWA}                     & 82.85                    & \multicolumn{1}{r|}{52.55} & 63.82                    & \multicolumn{1}{r|}{50.14}          & 68.45                    & \multicolumn{1}{r|}{50.73}          & \textbf{63.62}           & \textbf{50.00}           \\ \hline\hline
\end{tabular}}
\end{table}
\textbf{Results and Analysis.}
The results in Tables \ref{tab3} and \ref{tab5noniid} show that 
the defensive capability of WEF-Defense still achieves the SOTA performance 
when half or more than half of the clients are free-riders.
For instance, 
on all image datasets,
the HMA of global models obtained by free-riders is less than 36.50\%
when WEF-Defense is implemented.
On all structured datasets,
the HMA obtained by free-riders is less than 68.89\%
when WEF-Defense is conducted.
These are solid evidence of the stable defense effect of WEF-Defense at high free-rider ratios.

Meanwhile, 
WEF-Defense is more effective in preventing free-riders from obtaining high-quality models.
For instance,
compared with FedAvg on both tables,
the average HMA obtained by free-riders of WEF-Defense is reduced by 65\%,
while that of CFFL and RFFL is only reduced by 35\% and 40\%, respectively.
Besides,
WEF-Defense shows more stable performance than CFFL and RFFL 
when there are more free-riders than benign clients.
For instance,
in Table~\ref{tab3} with the IID setting,
the standard deviation of HMA for WEF-Defense on image datasets is around 9.85, 
while that for CFFL and RFFL reaches 25.74 and 26.29, respectively.
In Table~\ref{tab5noniid} with the Non-IID setting, 
the standard deviation of HMA for WEF-Defense is around 7.31, 
while that for CFFL and RFFL reaches 21.54 and 17.59, respectively.
The outstanding performance of WEF-Defense is mainly because
it identifies differences in the evolution process of the local model training, 
which effectively avoids model update camouflage for free-riders.

\begin{center}
\fcolorbox{black}{white!20}{\parbox{0.97\linewidth}
{
\emph{\textbf{Answer to RQ2:}}
When the number of free-riders is equal to or greater than that of benign clients, 
WEF-Defense shows better and more stable performance compared with baselines. 
Under the IID and Non-IID settings, on average,
1)~its defense effect is 1.41 and 1.28 times that of baselines, respectively; and 
2)~its defense stability is 2.64 and 2.67 times that of baselines, respectively.
}
}
\end{center}

\subsection{Defensive Timeliness}\label{sec4}
We conduct defense timeliness analysis for the experiments in RQ1 and RQ2,
where timeliness refers to earlier detection of free-riders.
Since CFFL cannot provide detection results during the training, 
we only compare the timeliness with RFFL.

\begin{table}[t]
\centering
\caption{For different free-rider ratios under the IID and Non-IID settings, the period
when the server confirms the free-riders during the total training rounds are recorded, respectively, 
where `\textbf{--}' represents that the defense method fails to detect the free-rider until the training ends,
`$t$ / $T$' represents that free-riders are detected in the $t$-th round when the total rounds is $T$.
}\label{table7}
\resizebox{1\textwidth}{!}{ \LARGE
\begin{tabular}{cccccccccccccccccc}
\hline\hline
                                   &                                                        & \multicolumn{8}{c}{\textbf{RFFL}}                                                                                                                                                                                                                                                               & \multicolumn{8}{c}{\textbf{WEF-Defense}}                                                                                                                                                                                                                                                       \\
\multirow{2}{*}{\textbf{Datasets}} & \multicolumn{1}{l|}{\multirow{2}{*}{\textbf{Attacks}}} & \multicolumn{4}{c|}{\textbf{Ratio under IID}}                                                                                                       & \multicolumn{4}{c|}{\textbf{Ratio under Non-IID}}                                                                                                   & \multicolumn{4}{c|}{\textbf{Ratio under IID}}                                                                                                       & \multicolumn{4}{c}{\textbf{Ratio under Non-IID}}                                                                                                   \\
                                   & \multicolumn{1}{l|}{}                                  & \multicolumn{1}{c}{\textbf{10\%}} & \multicolumn{1}{c}{\textbf{30\%}} & \multicolumn{1}{c}{\textbf{50\%}} & \multicolumn{1}{c|}{\textbf{90\%}} & \multicolumn{1}{c}{\textbf{10\%}} & \multicolumn{1}{c}{\textbf{30\%}} & \multicolumn{1}{c}{\textbf{50\%}} & \multicolumn{1}{c|}{\textbf{90\%}} & \multicolumn{1}{c}{\textbf{10\%}} & \multicolumn{1}{c}{\textbf{30\%}} & \multicolumn{1}{c}{\textbf{50\%}} & \multicolumn{1}{c|}{\textbf{90\%}} & \multicolumn{1}{c}{\textbf{10\%}} & \multicolumn{1}{c}{\textbf{30\%}} & \multicolumn{1}{c}{\textbf{50\%}} & \multicolumn{1}{c}{\textbf{90\%}} \\ \hline

\multirow{3}{*}{MNIST}    & \multicolumn{1}{c|}{RWA}              & 3/50          & 5/50          & \textbf{--}     & \multicolumn{1}{c|}{\textbf{--}}   & 1/50          & 5/50          & \textbf{--}   & \multicolumn{1}{c|}{\textbf{--}}   & \textbf{1/50} & \textbf{1/50} & \textbf{1/50} & \multicolumn{1}{c|}{\textbf{1/50}} & \textbf{1/50} & \textbf{1/50} & \textbf{1/50} & \textbf{1/50} \\
                          & \multicolumn{1}{c|}{SPA}              & 11/50         & 5/50          & 10/50           & \multicolumn{1}{c|}{21/50}         & 8/50          & 11/50         & 11/50         & \multicolumn{1}{c|}{\textbf{--}}   & \textbf{1/50} & \textbf{1/50} & \textbf{1/50} & \multicolumn{1}{c|}{\textbf{1/50}} & \textbf{1/50} & \textbf{1/50} & \textbf{1/50} & \textbf{1/50} \\
                          & \multicolumn{1}{c|}{DWA}              & \textbf{--}   & \textbf{--}   & \textbf{--}     & \multicolumn{1}{c|}{\textbf{--}}   & \textbf{--}   & \textbf{--}   & \textbf{--}   & \multicolumn{1}{c|}{\textbf{--}}   & \textbf{3/50} & \textbf{3/50} & \textbf{3/50} & \multicolumn{1}{c|}{\textbf{3/50}} & \textbf{3/50} & \textbf{3/50} & \textbf{3/50} & \textbf{3/50} \\ \hline
\multirow{3}{*}{CIFAR-10} & \multicolumn{1}{c|}{RWA}              & 10/80         & 11/80         & 16/80           & \multicolumn{1}{c|}{\textbf{--}}   & 11/80         & 20/80         & \textbf{--}           & \multicolumn{1}{c|}{\textbf{--}}   & \textbf{1/80} & \textbf{1/80} & \textbf{1/80} & \multicolumn{1}{c|}{\textbf{1/80}} & \textbf{1/80} & \textbf{1/80} & \textbf{1/80} & \textbf{1/80} \\
                          & \multicolumn{1}{c|}{SPA}              & 11/80         & 11/80         & 11/80           & \multicolumn{1}{c|}{\textbf{--}}   & 11/80         & 11/80         & 16/80         & \multicolumn{1}{c|}{\textbf{--}}   & \textbf{1/80} & \textbf{1/80} & \textbf{1/80} & \multicolumn{1}{c|}{\textbf{1/80}} & \textbf{1/80} & \textbf{1/80} & \textbf{1/80} & \textbf{1/80} \\
                          & \multicolumn{1}{c|}{DWA}              & \textbf{--}   & \textbf{--}   & \textbf{--}     & \multicolumn{1}{c|}{\textbf{--}}            & \textbf{--}   & \textbf{--}   & \textbf{--}   & \multicolumn{1}{c|}{\textbf{--}}            & \textbf{3/80} & \textbf{3/80} & \textbf{3/80} & \multicolumn{1}{c|}{\textbf{3/80}} & \textbf{3/80} & \textbf{3/80} & \textbf{3/80} & \textbf{3/80} \\ \hline
\multirow{3}{*}{GTSRB}    & \multicolumn{1}{c|}{RWA}              & 11/80         & 11/80         & \textbf{--}              & \multicolumn{1}{c|}{\textbf{--}}   & \textbf{--}            & \textbf{--}            & \textbf{--}            & \multicolumn{1}{c|}{\textbf{--}}   & \textbf{1/80} & \textbf{1/80} & \textbf{1/80} & \multicolumn{1}{c|}{\textbf{1/80}} & \textbf{1/80} & \textbf{1/80} & \textbf{1/80} & \textbf{1/80} \\
                          & \multicolumn{1}{c|}{SPA}              & 11/80         & 11/80         & 11/80           & \multicolumn{1}{c|}{\textbf{--}}   & 11/80         & 11/80         & 17/80         & \multicolumn{1}{c|}{\textbf{--}}   & \textbf{1/80} & \textbf{1/80} & \textbf{1/80} & \multicolumn{1}{c|}{\textbf{1/80}} & \textbf{1/80} & \textbf{1/80} & \textbf{1/80} & \textbf{1/80} \\
                          & \multicolumn{1}{c|}{DWA}              & \textbf{--}   & \textbf{--}   & \textbf{--}     & \multicolumn{1}{c|}{\textbf{--}}            & \textbf{--}   & \textbf{--}   & \textbf{--}   & \multicolumn{1}{c|}{\textbf{--}}   & \textbf{3/80} & \textbf{3/80} & \textbf{3/80} & \multicolumn{1}{c|}{\textbf{3/80}} & \textbf{3/80} & \textbf{3/80} & \textbf{3/80} & \textbf{3/80} \\ \hline
\multirow{3}{*}{ADULT}    & \multicolumn{1}{c|}{RWA}              & 19/50         & 10/50         & \textbf{--}     & \multicolumn{1}{c|}{\textbf{--}}   & \textbf{--}   & \textbf{--}   & \textbf{--}   & \multicolumn{1}{c|}{--}            & \textbf{1/50} & \textbf{1/50} & \textbf{1/50} & \multicolumn{1}{c|}{\textbf{1/50}} & \textbf{1/50} & \textbf{1/50} & \textbf{1/50} & \textbf{1/50} \\
                          & \multicolumn{1}{c|}{SPA}              & 10/50         & 10/50         & 10/50           & \multicolumn{1}{c|}{\textbf{--}}   & 11/50         & \textbf{--}            & 11/50         & \multicolumn{1}{c|}{21/80}         & \textbf{1/50} & \textbf{1/50} & \textbf{1/50} & \multicolumn{1}{c|}{\textbf{1/50}} & \textbf{1/50} & \textbf{1/50} & \textbf{1/50} & \textbf{1/50} \\
                          & \multicolumn{1}{c|}{DWA}              & \textbf{--}            & \textbf{--}            & \textbf{--}              & \multicolumn{1}{c|}{\textbf{--}}   & \textbf{--}\textbf{--}            & \textbf{--}            & \textbf{--}            & \multicolumn{1}{c|}{\textbf{--}}            & \textbf{3/50} & \textbf{3/50} & \textbf{3/50} & \multicolumn{1}{c|}{\textbf{3/50}} & \textbf{3/50} & \textbf{3/50} & \textbf{3/50} & \textbf{3/50} \\ \hline
\multirow{3}{*}{BANK}     & \multicolumn{1}{c|}{RWA}              & 2/80          & 3/80          & 5/80            & \multicolumn{1}{c|}{\textbf{--}}   & 11/80         & 11/80         & 10/80         & \multicolumn{1}{c|}{\textbf{--}}   & \textbf{1/80} & \textbf{1/80} & \textbf{1/80} & \multicolumn{1}{c|}{\textbf{1/80}} & \textbf{1/80} & \textbf{1/80} & \textbf{1/80} & \textbf{1/80} \\
                          & \multicolumn{1}{c|}{SPA}              & 10/80         & 5/80          & 7/80            & \multicolumn{1}{c|}{\textbf{--}}   & 11/80         & 11/80         & 8/80          & \multicolumn{1}{c|}{11/80}         & \textbf{1/80} & \textbf{1/80} & \textbf{1/80} & \multicolumn{1}{c|}{\textbf{1/80}} & \textbf{1/80} & \textbf{1/80} & \textbf{1/80} & \textbf{1/80} \\
                          & \multicolumn{1}{c|}{DWA}              & \textbf{--}            & \textbf{--}            &\textbf{--}              & \multicolumn{1}{c|}{\textbf{--}}   & \textbf{--}            & \textbf{--}            & \textbf{--}            & \multicolumn{1}{c|}{\textbf{--}}            & \textbf{3/80} & \textbf{3/80} & \textbf{3/80} & \multicolumn{1}{c|}{\textbf{3/80}} & \textbf{3/80} & \textbf{3/80} & \textbf{3/80} & \textbf{3/80} \\ \hline\hline
\end{tabular}}
\end{table}

\begin{table}[!htpb]\label{Significance}
\centering
\caption{The p-value of T-test.}
\resizebox{0.75\textwidth}{!}{%
\begin{tabular}{ccrrrr}
\hline\hline
\multicolumn{1}{l}{\multirow{2}{*}{\textbf{Method Comparison}}} & \multicolumn{1}{l}{\multirow{2}{*}{\textbf{Distributions}}} & \multicolumn{4}{c}{\textbf{Ratios}}                           \\
\multicolumn{1}{l}{}                                            & \multicolumn{1}{l}{}                                       & \textbf{10\%} & \textbf{30\%} & \textbf{50\%} & \textbf{90\%} \\ \hline
\multirow{2}{*}{FadAvg---WEF-Defense}     & IID        & 1.59E-05  & 2.84E-04  & 4.63E-04 &6.42E-04
               \\
& Non-IID    &1.59E-05   & 3.71E-04   &1.69E-04  &1.83E-03
                  \\ \cline{2-6}
\multirow{2}{*}{CFFL---WEF-Defense}       & IID        &4.52E-04
                  &1.38E-03
                & 2.33E-03
                 & 4.53E-02
                 \\
                                & Non-IID    & 8.66E-03
                & 9.82E-03
             &1.03E-02
                &2.30E-02
                 \\ \cline{2-6}
\multirow{2}{*}{RFFL---WEF-Defense}       & IID        &6.96E-03
                 &1.90E-02
                 &1.48E-02
               &2.55E-02
                  \\
                                & Non-IID    &6.32E-02
                &1.93E-02
                &3.58E-02
               &2.04E-02
               \\ \hline\hline
\end{tabular}}\label{Significance}
\end{table}

As shown in Table \ref{table7},
WEF-Defense is capable of free-rider detection at an earlier period compared with RFFL on all datasets. 
For instance, 
for almost all cases, 
WEF-Defense detects free-riders in the first round, 
while RFFL fails to detect them until the end of training in most cases.
The reason is that based on the collected WEF-Matrix information, 
it can distinguish free-riders and benign clients easily. 
Besides, 
it is difficult for free-riders to disguise WEF-Matrix,
so WEF-Defense can identify free-riders earlier.

\subsection{Significance Analysis}\label{sec4}
To illustrate the superiority of WEF-Defense's effect, 
we perform a preliminary T-test for the experiments in RQ1 and RQ2, 
compared with baselines, 
to confirm whether there is a significant difference in the defense effect of WEF-Defense. 
The results are shown in Table \ref{Significance}.

For the T-test, we define the null hypothesis as that the differences between defense methods are small. 
From the experimental results, 
we can see that the overall p-value is small enough ($<$0.05) to reject the null hypothesis, which proves the superiority of WEF-Defense.

\subsection{RQ3: Trade-off between Defense and Main Task Performance}\label{sec4}
In this section, we discuss whether defensive approaches sacrifice main task performance.


\begin{figure}[]
\centering
    \subfigure[Random weight attack (free-riders=10\%) ]{
    \includegraphics[width=0.47\linewidth]{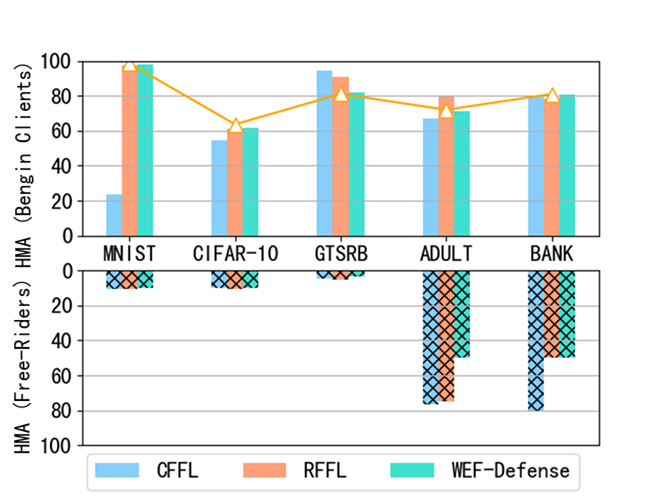}}
    \subfigure[\fm{Random weight attack (free-riders=90\%)}]{
    \includegraphics[width=0.47\linewidth]{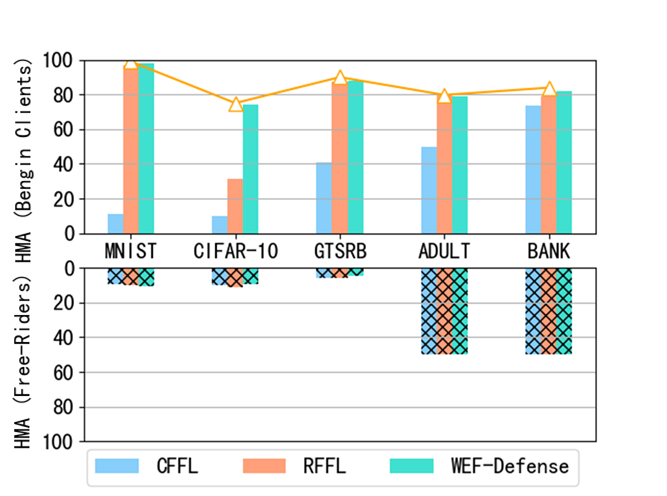}}\\
    \subfigure[Stochastic perturbations attack (free-riders=10\%)]{
    \includegraphics[width=0.47\linewidth]{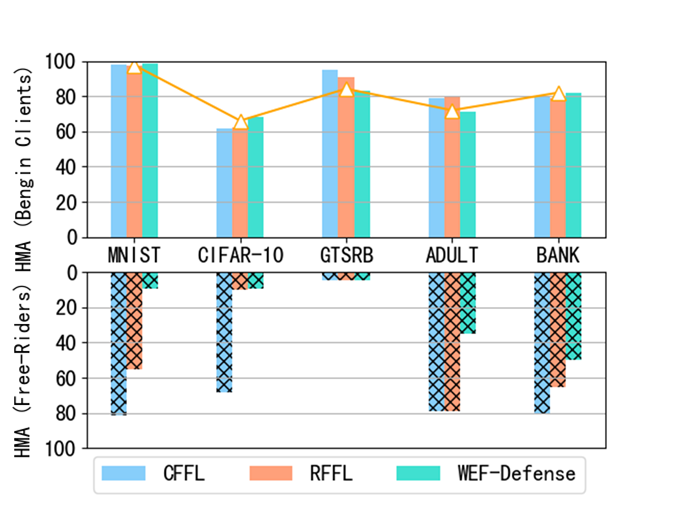}}
    \subfigure[\fm{Stochastic perturbations attack (free-riders=90\%)}]{
    \includegraphics[width=0.47\linewidth]{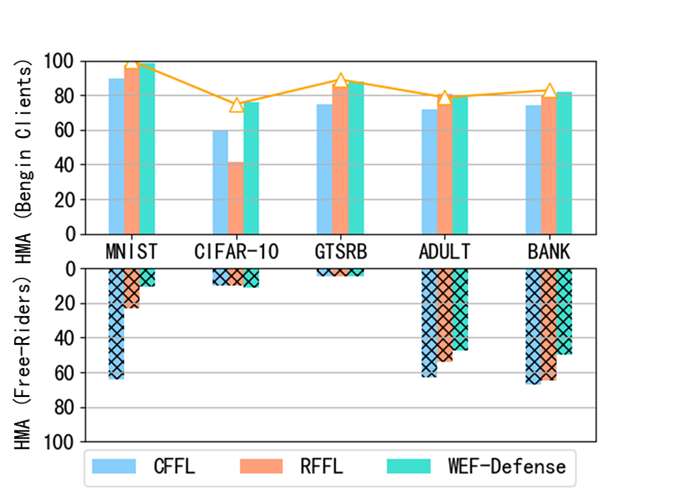}}\\
     \subfigure[\fm{Delta weight attack (free-riders=10\%)}]{
    \includegraphics[width=0.47\linewidth]{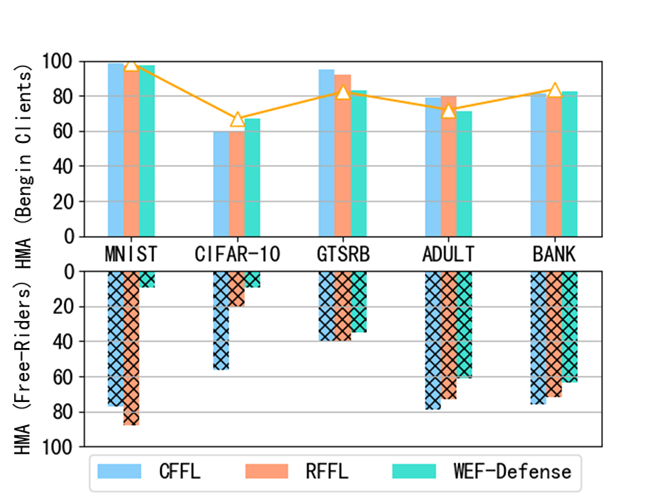}}
    \subfigure[\fm{Delta weight attack (free-riders=90\%)}]{
    \includegraphics[width=0.47\linewidth]{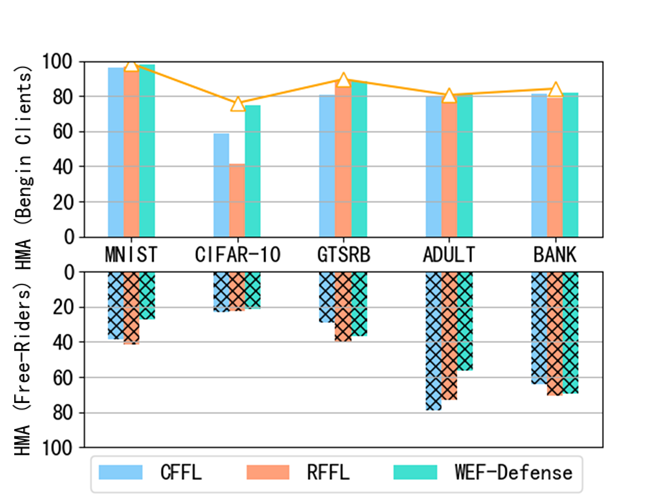}}
   \caption{HMA comparision between benign clients and free-riders. 
   The line represents HMA obtained by the benign clients in the no free-rider scenario.
   }\label{shuang}
\end{figure}


\subsubsection{Comparison of Global Model Accuracy}
\textbf{Implementation Details.}
(1)~Since CFFL and RFFL are contribution-based federated frameworks, 
we only consider testing under the IID setting. 
(2)~We test two scenarios of 10\% and 90\% free-riders in 10 clients. 
We plot HMA obtained by benign clients and free-riders respectively, 
then compare HMA obtained by benign clients in the no free-rider scenario, 
as shown in Fig.\ref{shuang}.

\textbf{Results and Analysis.}
Experiments show that WEF-Defense can not only defend against free-rider attacks, 
but also maintains the high accuracy of global models by aggregating all benign clients via eliminating the negative effect of free-riders' updates.
In Fig.\ref{shuang},
it is more significant in the scene where the free-rider accounts for 90\%. 
For instance,
on CIFAR-10, 
the overall average HMA obtained by benign clients with RFFL (44.04\%) and CFFL (51.73\%)
is much lower than that with WEF-Defense (78.13\%).

Comparing subfigures (a) and (b) in Fig.\ref{shuang}, 
we notice that random weight attack decreases the global model accuracy as the number of free-riders increases. 
Benefiting from the personalized aggregation,
WEF-Defense eliminates the impact of the random weight by leaving out the update from the free-riders.
Therefore,
the trained model for benign clients achieves the trade-off between accuracy and defensibility.


Observing the lines in Fig.\ref{shuang},
we can conclude that HMA of global model trained with only benign clients and trained with WEF-Defense are close,
where ``$\triangle$'' represents HMA trained with only benign clients. 
Comparing the main performance of different defense methods, 
especially on the CIFAR-10 dataset with the free-rider ratio of 90\%, the main performance of the global model with baselines is affected,
and only WEF-Defense achieves the expected primary performance.
It is mainly attributed to WEF-Defense that separates free-riders and benign clients into groups $\left \{ P_n, P_r \right \}$, 
and adopts a personalized federated learning approach to provide them with different global models.

\subsubsection{Time Complexity Analysis}

Compared with the FedAvg, WEF-Defense requires each client to upload not only the updated weights of the local model, but also the WEF-Matrix for free-rider detection. 
Thus the communication overhead of WEF-Defense is calculated to analyze its complexity.

\begin{table}[h]
\centering
\caption{Time cost (second) comparison betweent WEF-Defense and FedAvg.}\label{time}
\resizebox{0.7\textwidth}{!}{%
\begin{tabular}{lllrrrr}
\hline\hline
\multicolumn{1}{c}{\multirow{2}{*}{\textbf{Datasets}}} & \multicolumn{1}{c}{\multirow{2}{*}{\textbf{Models}}} & \multicolumn{1}{c}{\multirow{2}{*}{\textbf{Methods}}} & \multicolumn{4}{c}{\textbf{Ratios}}                                                                                                            \\
\multicolumn{1}{c}{}                                  & \multicolumn{1}{c}{}                                & \multicolumn{1}{c}{}                                 & \multicolumn{1}{c}{\textbf{10\%}} & \multicolumn{1}{c}{\textbf{30\%}} & \multicolumn{1}{c}{\textbf{50\%}} & \multicolumn{1}{c}{\textbf{90\%}} \\ \hline
\multirow{2}{*}{MNIST}                                & \multirow{2}{*}{LeNet}                              & FedAvg                                               & 28.71                             & 22.35                             & 16.82                             & 6.51                               \\
                                                      &                                                     & WEF-Defense                                          & 29.97                             & 24.55                             & 17.82                             & 6.88                              \\ \hline
\multirow{2}{*}{CIFAR-10}                             & \multirow{2}{*}{VGG16}                              & FedAvg                                               & 204.98                            & 170.38                            & 133.51                            & 54.85                             \\
                                                      &                                                     & WEF-Defense                                          & 211.98                            & 180.35                            & 167.64                            & 78.52                             \\ \hline
\multirow{2}{*}{GTSRB}                                & \multirow{2}{*}{ResNet18}                           & FedAvg                                               & 284.74                            & 222.35                            & 159.57                            & 80.38                             \\
                                                      &                                                     & WEF-Defense                                          & 318.23                            & 252.82                            & 184.79                            & 85.82                              \\ \hline
\multirow{2}{*}{ADULT}                                & \multirow{2}{*}{MLP}                                & FedAvg                                               & 7.15                              & 5.01                              & 3.77                              & 1.62                              \\
                                                      &                                                     & WEF-Defense                                          & 9.19                              & 5.82                              & 4.42                              & 2.22                              \\ \hline
\multirow{2}{*}{BANK}                                 & \multirow{2}{*}{MLP}                                & FedAvg                                               & 2.84                              & 1.78                              & 1.54                              & 0.61                              \\
                                                      &                                                     & WEF-Defense                                          & 4.33                              & 3.11                              & 2.53                              & 1.08                              \\ \hline\hline
\end{tabular}}
\end{table}

\textbf{Implementation Details.} 
We evaluate and compare the time cost of one epoch for normal training (i.e., FedAvg) and WEF-Defense training on each dataset.
The experimental results are shown in Table \ref{time}.

\textbf{Results and Analysis.}
The time cost of WEF-Defense is tolerable.
For instance, on CIFAR-10 and GTSRB in Table \ref{time}, 
WEF-Defense takes about 3.41\% to 43.15\% more time than the normal training process. 
Moreover, several datasets (e.g., ADULT and BANK) are easier to train, 
thus their time cost is negligible.

\begin{center}
\fcolorbox{black}{white!20}{\parbox{0.97\linewidth}
{
\emph{\textbf{Answer to RQ3:}}
In summary, WEF-Defense can trade-off defense and main task performance.
On average, 
1)~WEF-Defense outperforms CFFL and RFFL by 33.02\% and 10.73\% on the main task performance when facing high-proportion free-riders, respectively;
2)~compared with FedAvg, 
WEF-Defense only takes an average of 23.8\% more time.
}
}
\end{center}



\subsection{RQ4: Interpretation of WEF-Defense via Visualization}
We further illustrate the effectiveness of WEF-Defense by visualization.


\begin{figure}[]
\centering
    \subfigure[The variation process of the global models' accuracy when the free-rider ratio is 10\%]{
    \includegraphics[width=0.3\linewidth]{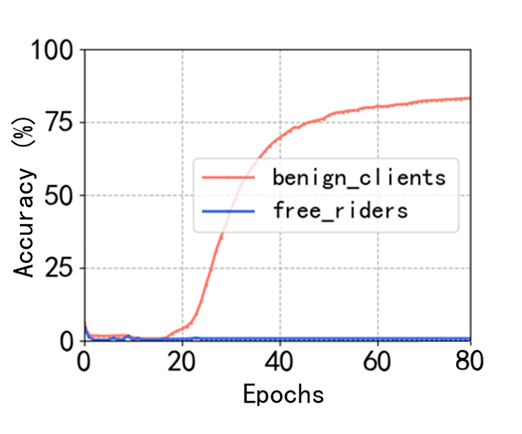}
    \includegraphics[width=0.3\linewidth]{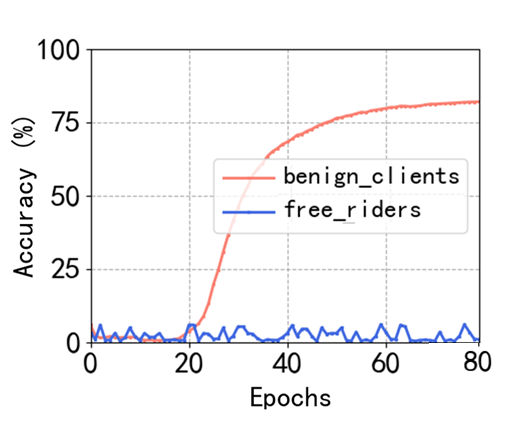}
    \includegraphics[width=0.3\linewidth]{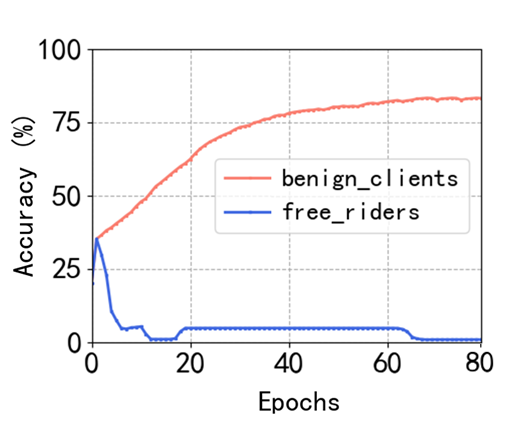}    } \\
     \subfigure[The variation process of the global models' accuracy when the free-rider ratio is 30\%]{
    \includegraphics[width=0.3\linewidth]{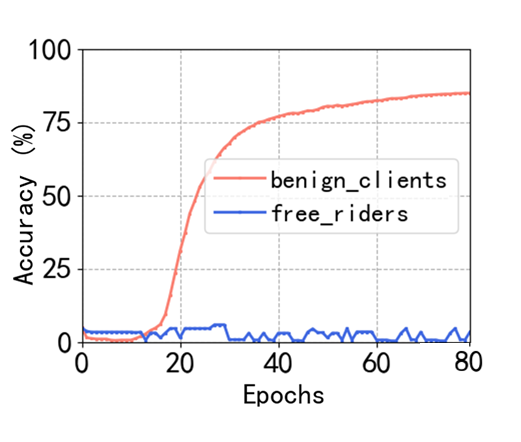}
    \includegraphics[width=0.3\linewidth]{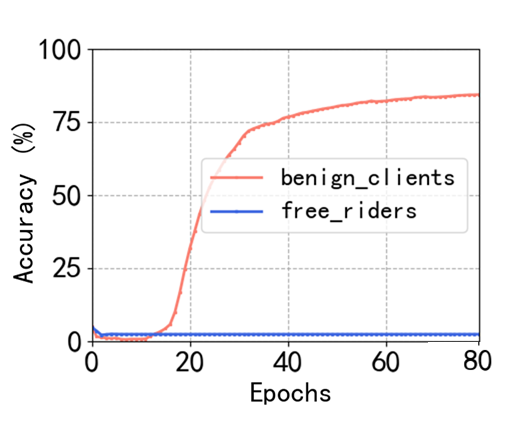}
    \includegraphics[width=0.3\linewidth]{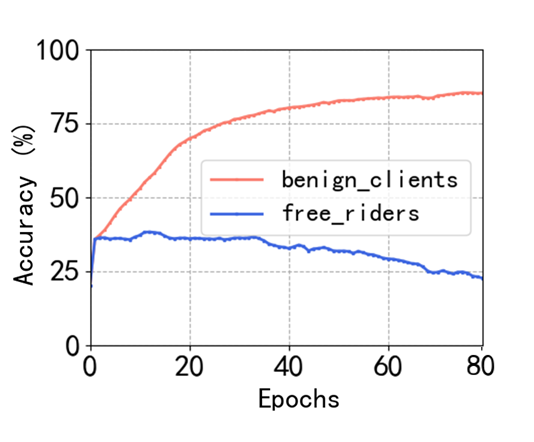}}\\
     \subfigure[The variation process of the global models' accuracy when the free-rider ratio is 50\%]{
    \includegraphics[width=0.3\linewidth]{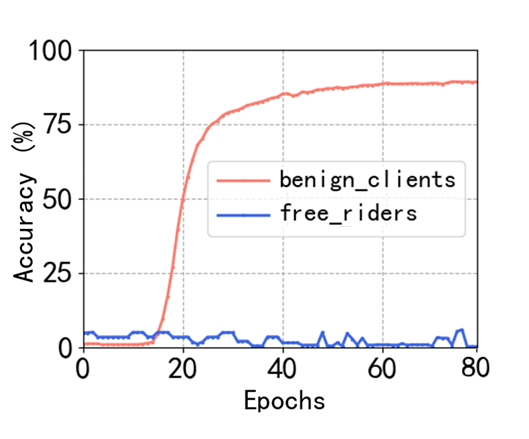}
    \includegraphics[width=0.3\linewidth]{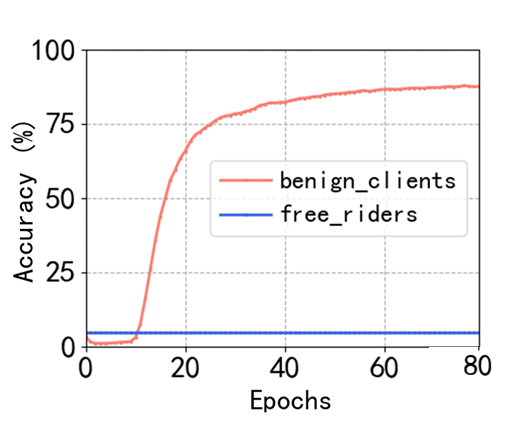}
    \includegraphics[width=0.3\linewidth]{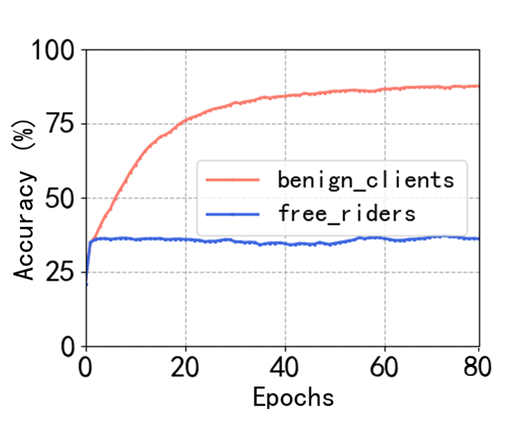}}\\
     \subfigure[The variation process of the global models' accuracy when the free-rider ratio is 90\%]{
    \includegraphics[width=0.3\linewidth]{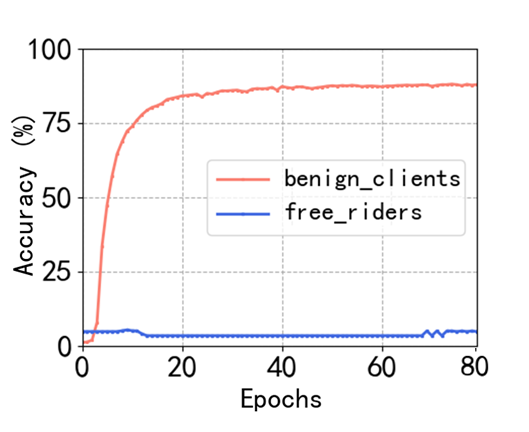}
    \includegraphics[width=0.3\linewidth]{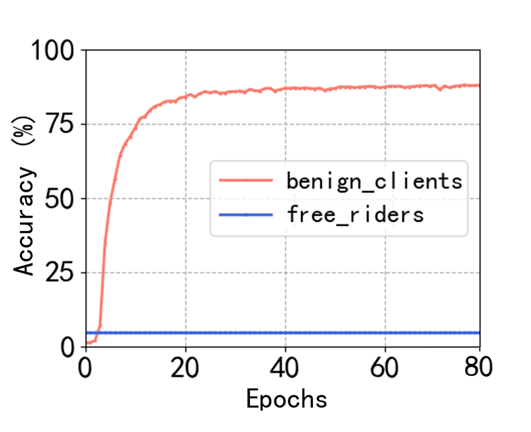}
    \includegraphics[width=0.3\linewidth]{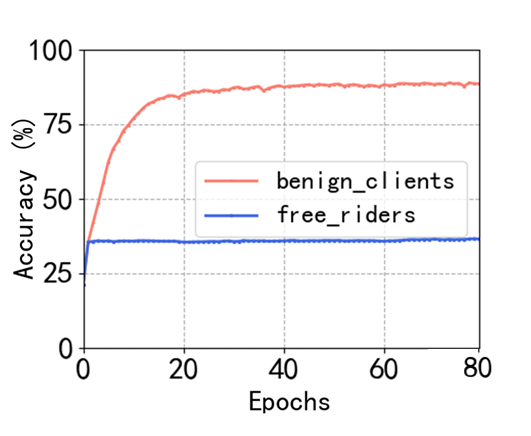}}
   \caption{
The process of global models' accuracy variation obtained by benign clients and free-riders during personalized federation training in the GTSRB dataset under the IID setting.
For each subfigure, from left to right, experimental results of WEF-Defense against random weight attack, stochastic perturbations attack and delta weight attack are shown.
   }\label{fig10}
\end{figure}


\begin{figure}[]
\centering
    \subfigure[WEF-Matrix of bengin clients on GTSRB dataset]{
    \includegraphics[width=1\linewidth]{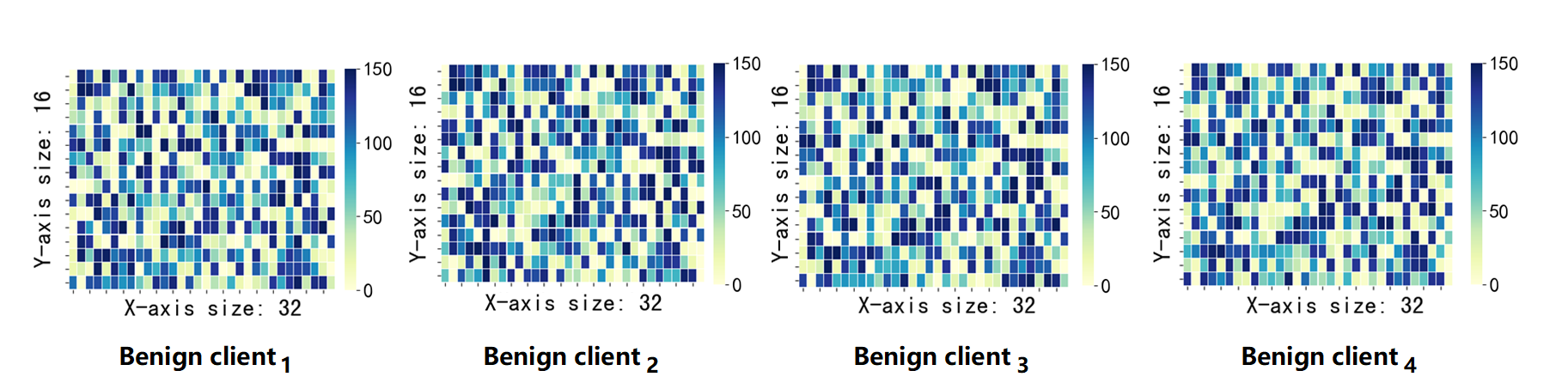}}\\
    \subfigure[\fm{ WEF-Matrix of free-riders on GTSRB dataset}]{
    \includegraphics[width=1\linewidth]{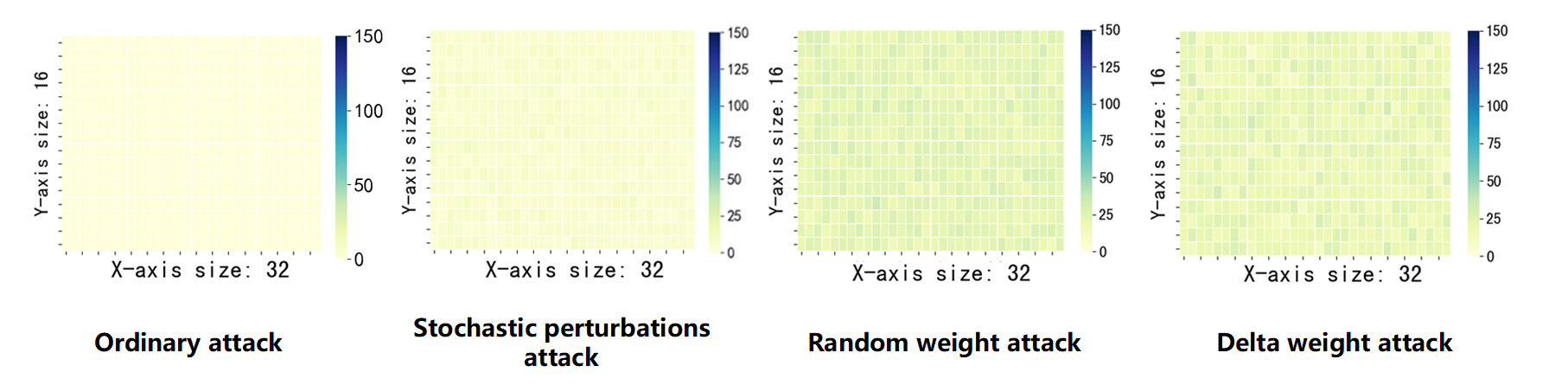}}\\
    \subfigure[ WEF-Matrix of bengin clients on CIFAR-10 dataset]{
    \includegraphics[width=1\linewidth]{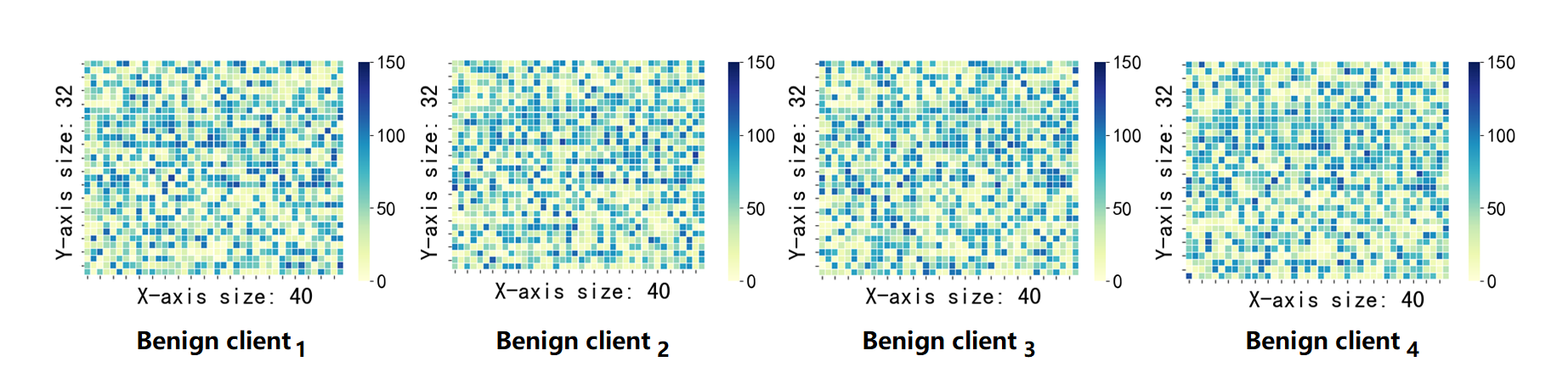}} \\
    \subfigure[\fm{ WEF-Matrix of free-riders on CIFAR-10 dataset}]{
    \includegraphics[width=1\linewidth]{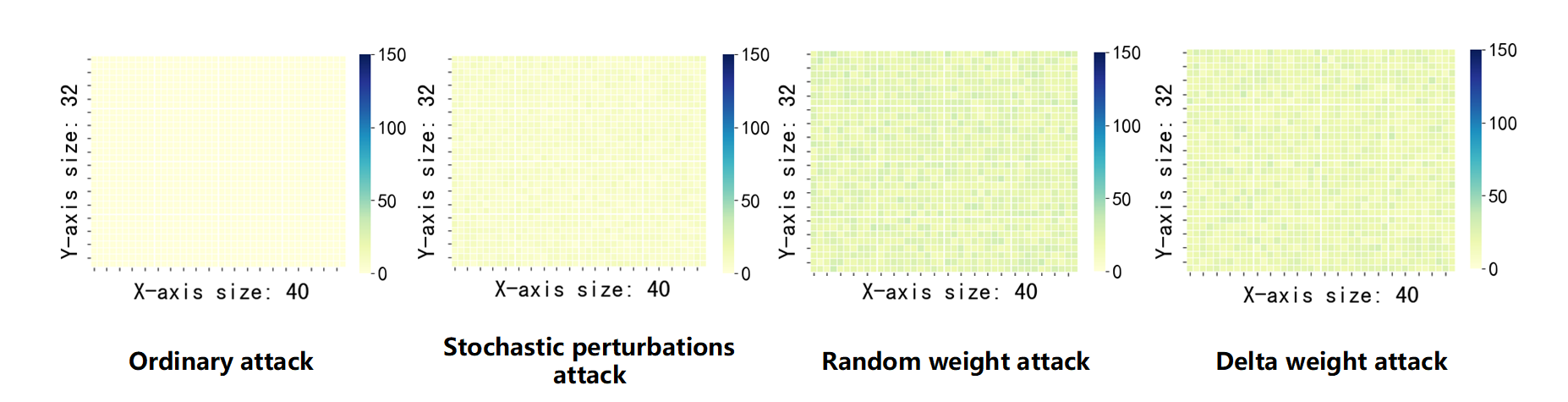}}
   \caption{WEF-Matrix visualization of benign clients and free-riders on different datasets.}\label{fig11}
\end{figure}

\textbf{Implementation Details.}
(1) For federated training with WEF-Defense, 
we visualize the global model accuracy obtained by two groups $\left \{ P_n, P_r \right \}$ on the GTSRB dataset.
The results are shown in Fig.\ref{fig10}.
(2) In the experiment under the IID setting, 
WEF-Matrix of the four selected benign clients and four free-rider attacks 
(including the original free-rider attack) 
are displayed in the heatmap, 
as shown in Fig.\ref{fig11}.
More visualization results we put in the \ref{appendix_visualization}.

\textbf{Results and Analysis.}
At the early stage of federated training, 
the server can completely separate benign clients and free-riders, 
as shown in Fig.\ref{fig10}.
Consequently, 
WEF-Defense is capable of preventing free-riders to obtain a high-quality model.
Meanwhile, after free-riders are separated from the benign clients, 
the accuracy of global models assigned to free-riders is low or even degraded,
while benign clients can train collaboratively to build high-quality models.

The superiority of defense timeliness is because WEF-Matrix can effectively distinguish benign clients from free-riders.
It is obvious from Fig.\ref{fig11} that, on the one hand, the model weight evolving frequency of benign clients has a certain evolving pattern, e.g., some weights evolving frequency are much larger than others, indicating that during normal training, the input data has a greater impact on the weights and has strong activation.
Some weights do not have a large frequency variation, indicating that some neurons are difficult to activate, resulting in a weaker optimization of the weights.
On the other hand, in the free-rider's WEF-Matrix, the original free-rider attack does not perform any operation on the model issued by the server, so the weight does not have any optimization process, resulting in the overall weight variation frequency of 0.

In the other three free-rider attacks, although different degrees of camouflage are used, it is difficult to identify sensitive and insensitive neurons because the local model did not carry out normal training. Meanwhile, 
due to the non-sharing between clients, stealing the optimization results of each weight is difficult. 
Therefore, it is a challenge for camouflage methods to correctly simulate the variation frequency of each weight, which leads to a very large difference from the WEF-Matrix of the benign client.
This enables the server to separate free-riders from benign clients in the early stages of training.

\begin{center}
\fcolorbox{black}{white!20}{\parbox{0.97\linewidth}
{
\emph{\textbf{Answer to RQ4:}}
Through the visual analysis of WEF-Defense, 
the effectiveness of WEF-Matrix is further illustrated,
it outperforms baselines in two aspects: 
(1)~it prevents free-riders from acquiring the global model contributed by benign clients at the early stage of training; 
(2)~it does not affect the global model trained by benign clients.
}
}
\end{center}



\subsection{RQ5: Defense against Adaptive Attacks and Hyperparameter Sensitivity Analysis}\label{rq5}
This section discusses the effectiveness of WEF-Defense system against adaptive attack and the sensitivity analysis of hyperparameter to different experimental settings.
\subsubsection{Defense against Adaptive Attack}\label{ada}
When answering this question, 
we design an adaptive attack method to test the robustness of WEF-Defense. 
The attack absorbs the advantages of the most camouflaged delta weight attack. Meanwhile, to imitate the evolving frequency of the weight during the training process of the benign client, so as to narrow the difference with the WEF-Matrix of the benign client.

\textbf{Implementation Details.}
(1) For our proposed WEF-Defense, we design an adaptive free-rider attack method that simulates the normal training and weight optimization process of benign clients.
It adds different perturbation noise in the local training round, including $N(0, 10^{-3})$, $N(0, 10^{ -4}) $, $N(0, 10^{ -5})$ normally distributed noise.
(2) Experiments are conducted under the IID and Non-IID data of five datasets. The results are shown in the Table \ref{tab5}.

\begin{table}[thb]
\centering
\caption{
HMA (\%) that can be obtained by free-riders (\%) when WEF-Defense faces an adaptive attack.
The experiments are tested under the IID and Non-IID settings with different free-rider ratios.
}\label{tab5}
\resizebox{0.7\textwidth}{!}{%
\begin{tabular}{ccccccccc}
\hline\hline
\textbf{}                             & \multicolumn{4}{c}{\textbf{Ratio under IID}}                                                   & \multicolumn{4}{c}{\textbf{Ratio under Non-IID}}                          \\
\multicolumn{1}{c|}{\textbf{Datasets}} & \textbf{10\%} & \textbf{30\%} & \textbf{50\%} & \multicolumn{1}{c|}{\textbf{90\%}} & \textbf{10\%} & \textbf{30\%} & \textbf{50\%} & \textbf{90\%} \\ \hline
\multicolumn{1}{c|}{MNIST}           & 10.14 & 20.12 & 19.71 & \multicolumn{1}{c|}{24.2}  & 21.74 & 15.63 & 21.52 & 23.41          \\
\multicolumn{1}{c|}{CIFAR-10}         & 10.15 & 21.53 & 20.52 & \multicolumn{1}{c|}{20.45} & 9.47  & 18.31 & 18.66 & 13.63          \\
\multicolumn{1}{c|}{GTSRB}           & 35.27 & 36.46 & 36.22 & \multicolumn{1}{c|}{35.95} & 30.04 & 26.99 & 34.71 & 23.04 \\
\multicolumn{1}{c|}{ADULT}             & 64.65 & 56.66 & 56.36 & \multicolumn{1}{c|}{55.94} & 63.63 & 51.82 & 51.73 & 49.96 \\
\multicolumn{1}{c|}{BANK}             & 62.55 & 70.05 & 69.55 & \multicolumn{1}{c|}{68.95} & 67.45 & 50.00 & 63.67 & 50.00           \\ \hline\hline
\end{tabular}}
\end{table}

\textbf{Results and Analysis.}
We find that the designed adaptive attack is still difficult to obtain a high-quality model under WEF-Defense as shown in Table \ref{tab5}.
Although we try to simulate the weights optimization process of benign clients by adding different noises.
Since the free-rider itself is not trained, and the information of the local training of the benign client cannot be obtained. 
It is difficult for the camouflage methods to simulate the optimization process correctly, 
resulting in a difference from the WEF-Matrix of benign clients.

\begin{figure}[]
\centering
    \subfigure[\fm{IID 10\%-attack} ]{
    \includegraphics[width=0.35\linewidth]{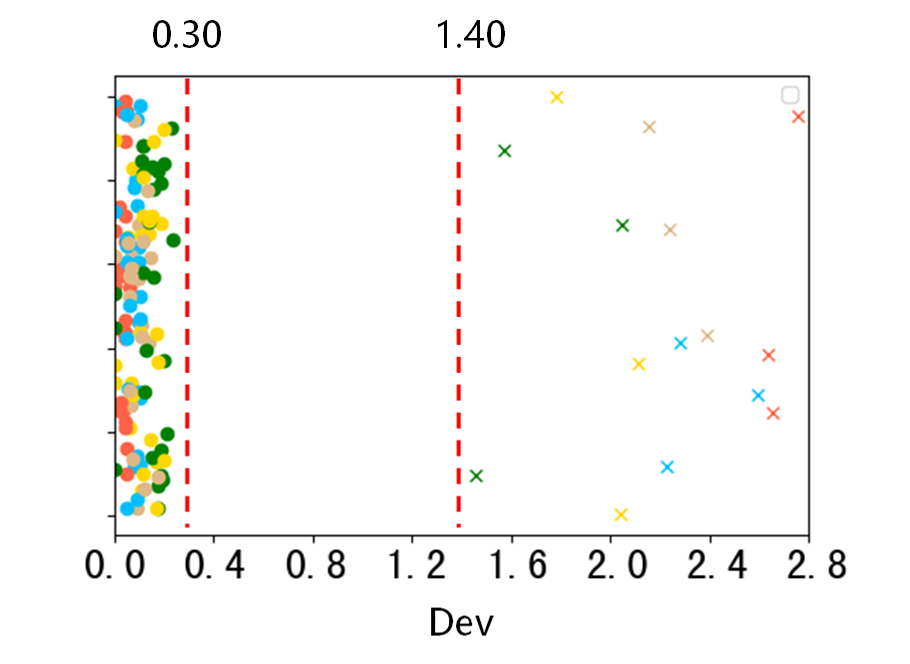}}
    \subfigure[\fm{Non-IID 10\%-attack}]{
    \includegraphics[width=0.35\linewidth]{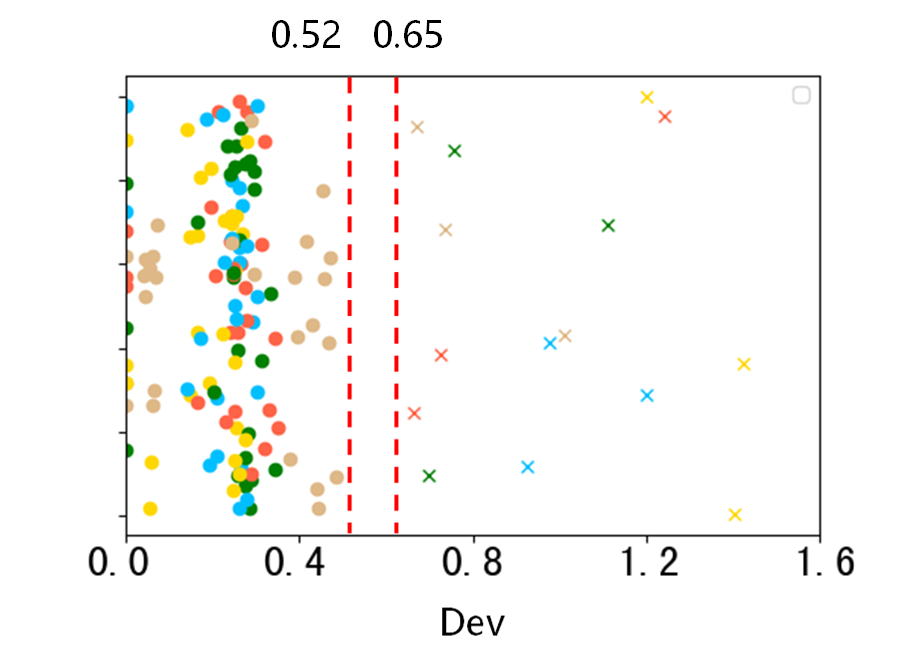}}\\
    \subfigure[\fm{IID 30\%-attack}]{
    \includegraphics[width=0.35\linewidth]{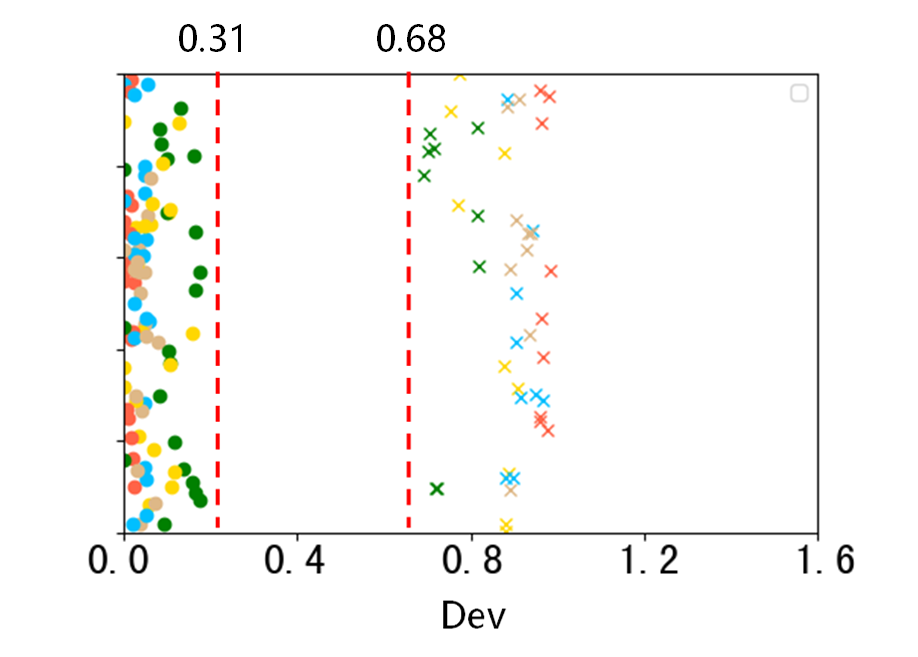}}
    \subfigure[\fm{Non-IID 30\%-attack}]{
    \includegraphics[width=0.35\linewidth]{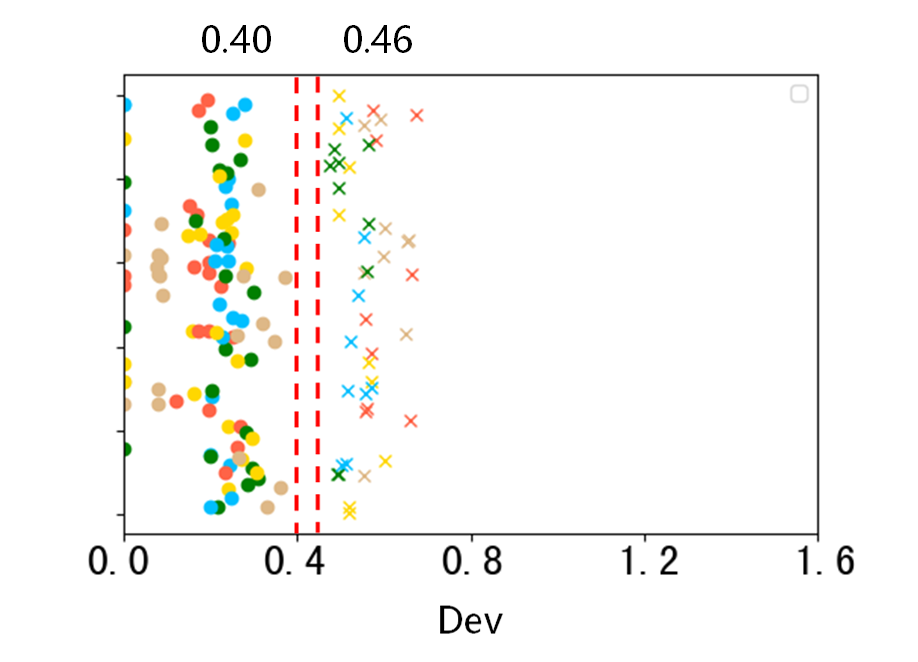}}\\
     \subfigure[\fm{IID 50\%-attack}]{
    \includegraphics[width=0.35\linewidth]{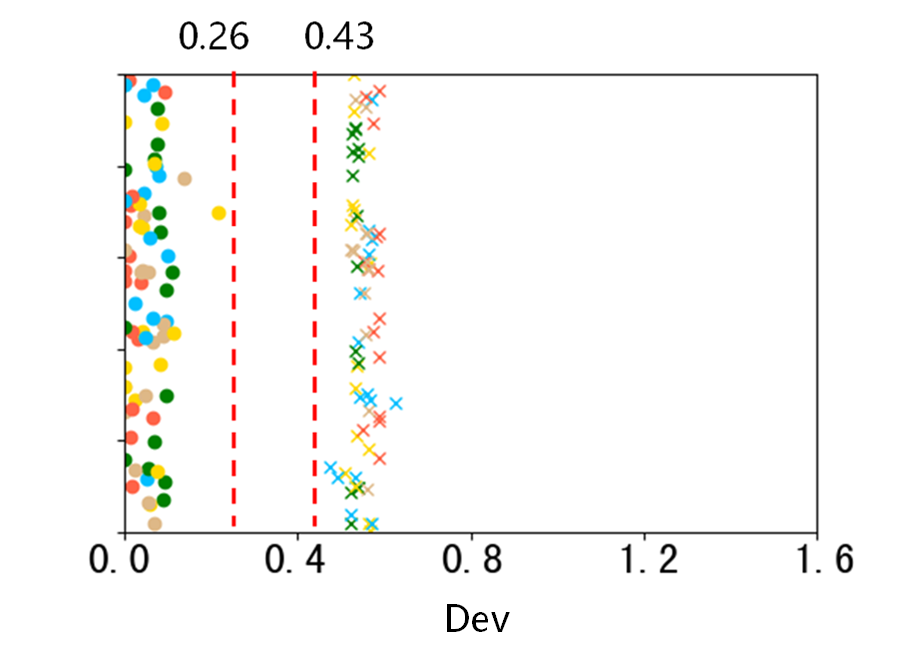}}
    \subfigure[\fm{Non-IID 50\%-attack}]{
    \includegraphics[width=0.35\linewidth]{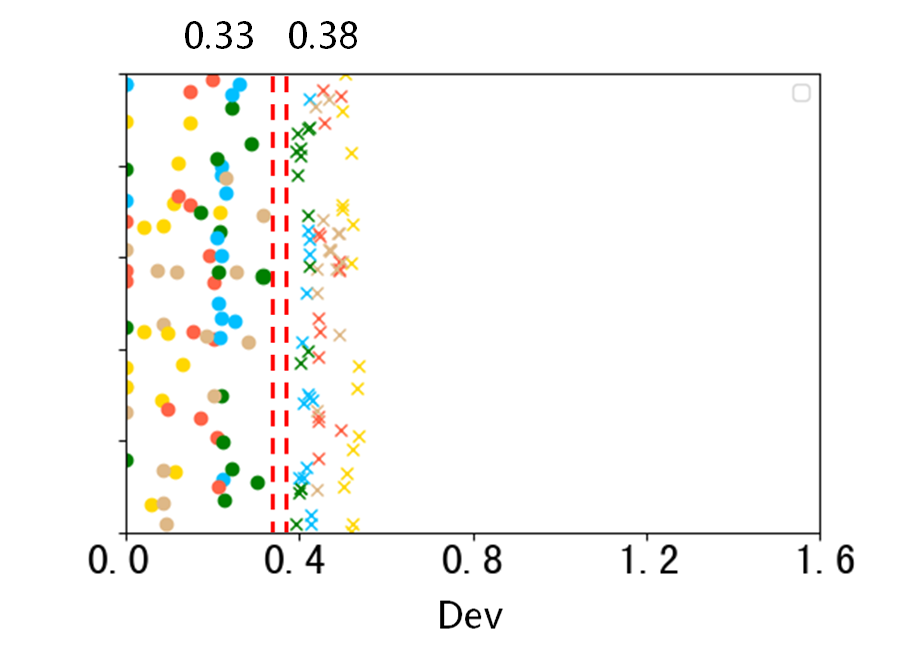}} \\
    \subfigure[\fm{IID 90\%-attack}]{
    \includegraphics[width=0.35\linewidth]{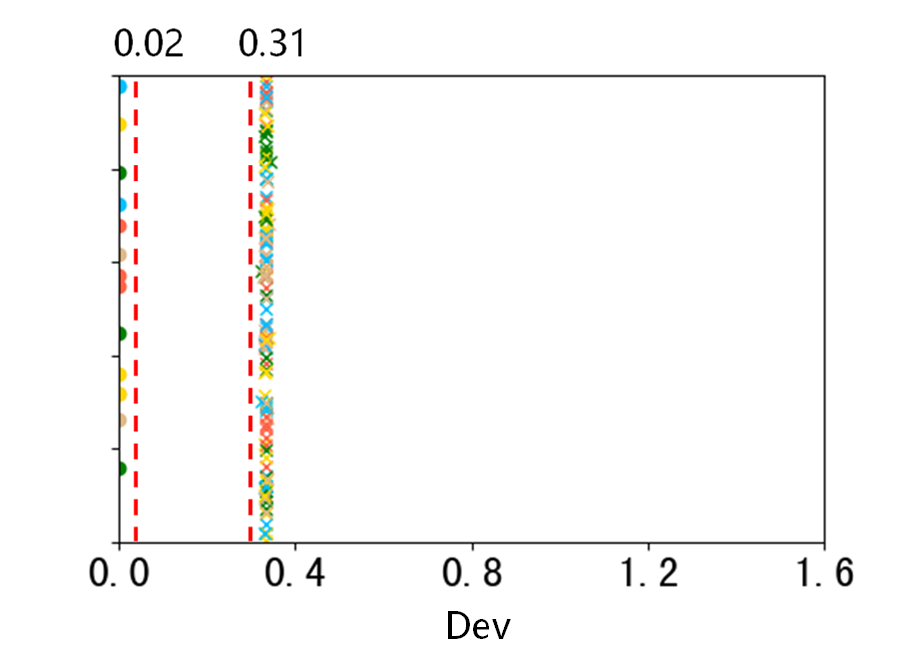}}
    \subfigure[\fm{Non-IID 90\%-attack}]{
    \includegraphics[width=0.35\linewidth]{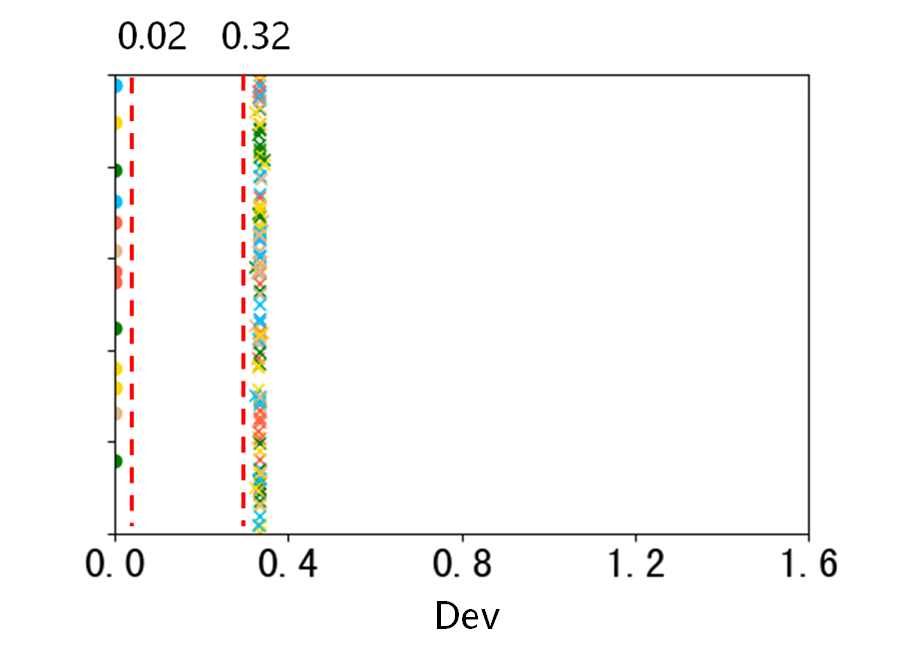}} \\
    \centering
    \includegraphics[width=0.7\linewidth]{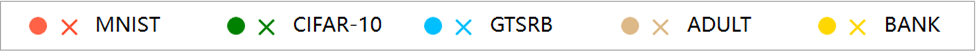}
   \caption{
The similarity deviation values $Dev$ for benign clients and free-riders are visualized, including five datasets under the IID and Non-IID experimental settings.
We make a unified analysis under the condition that the rate of free-rider is the same,
where ``$\bullet$'' and ``$\times$''  represent the benign client and free-rider, respectively.
}\label{canshu10}
\end{figure}

\begin{figure}[]
\centering
    \subfigure[\fm{IID 10\%-attack} ]{
    \includegraphics[width=0.35\linewidth]{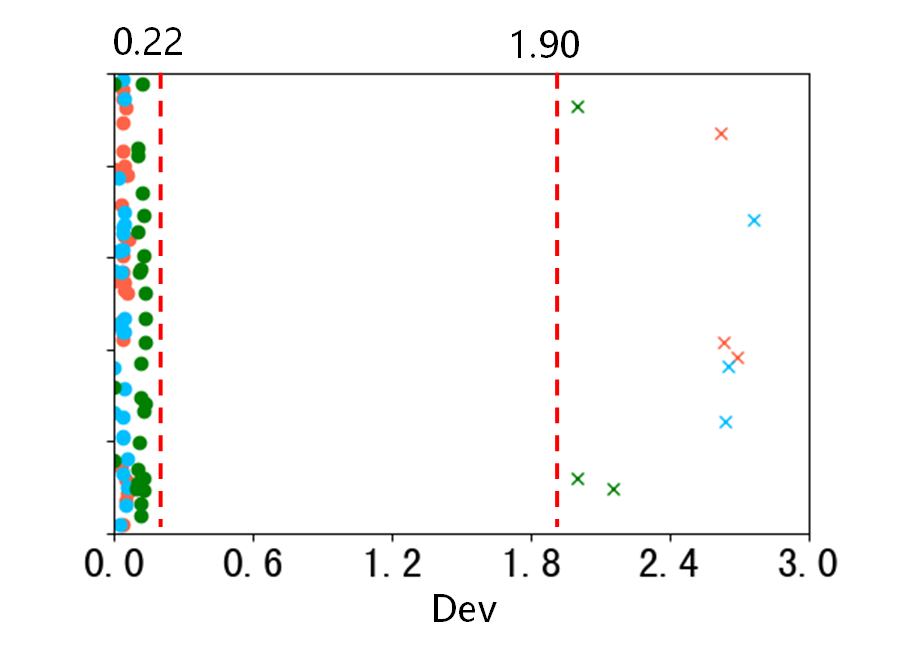}}
    \subfigure[\fm{Non-IID 10\%-attack}]{
    \includegraphics[width=0.35\linewidth]{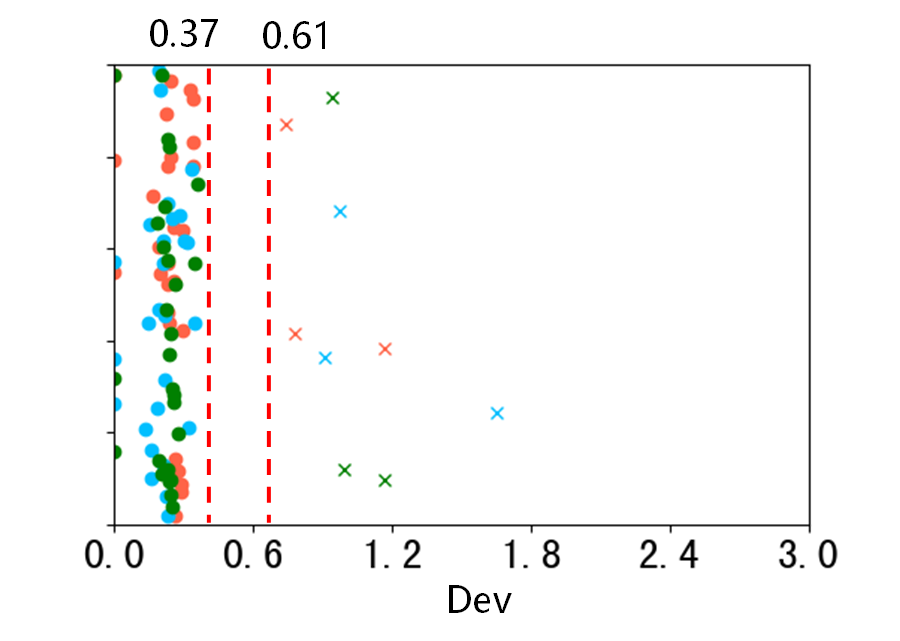}}\\
    \subfigure[\fm{IID 30\%-attack}]{
    \includegraphics[width=0.35\linewidth]{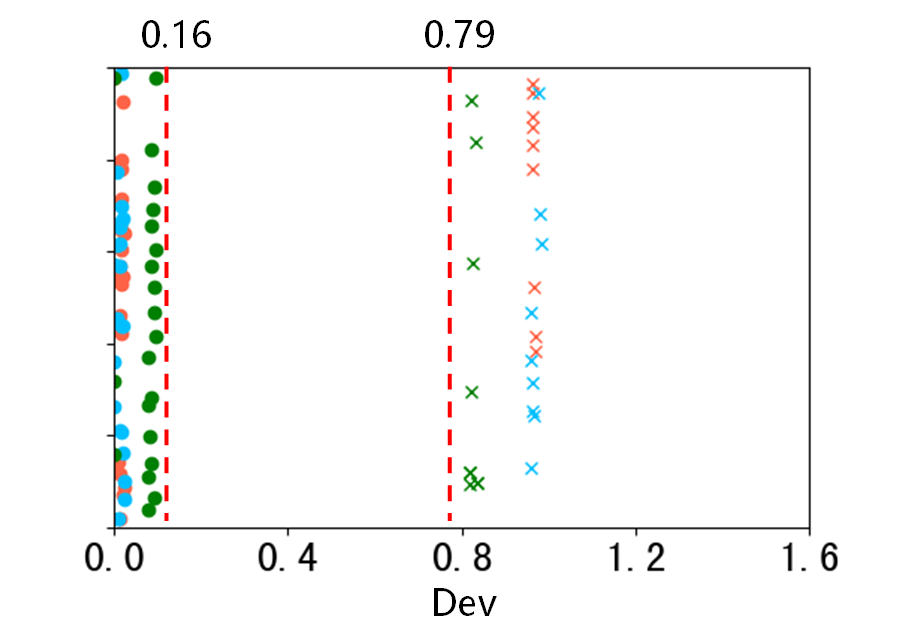}}
    \subfigure[\fm{Non-IID 30\%-attack}]{
    \includegraphics[width=0.35\linewidth]{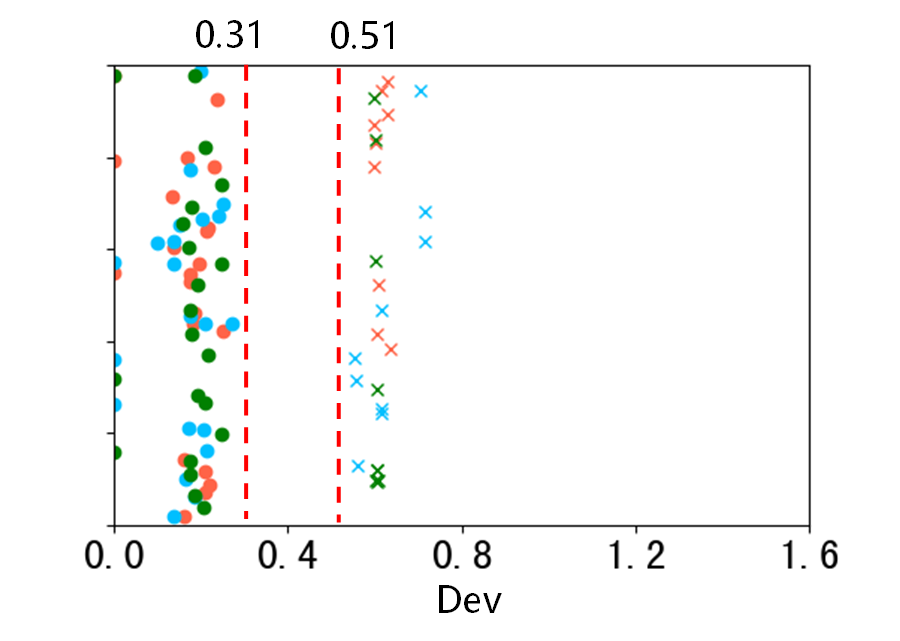}}\\
     \subfigure[\fm{IID 50\%-attack}]{
    \includegraphics[width=0.35\linewidth]{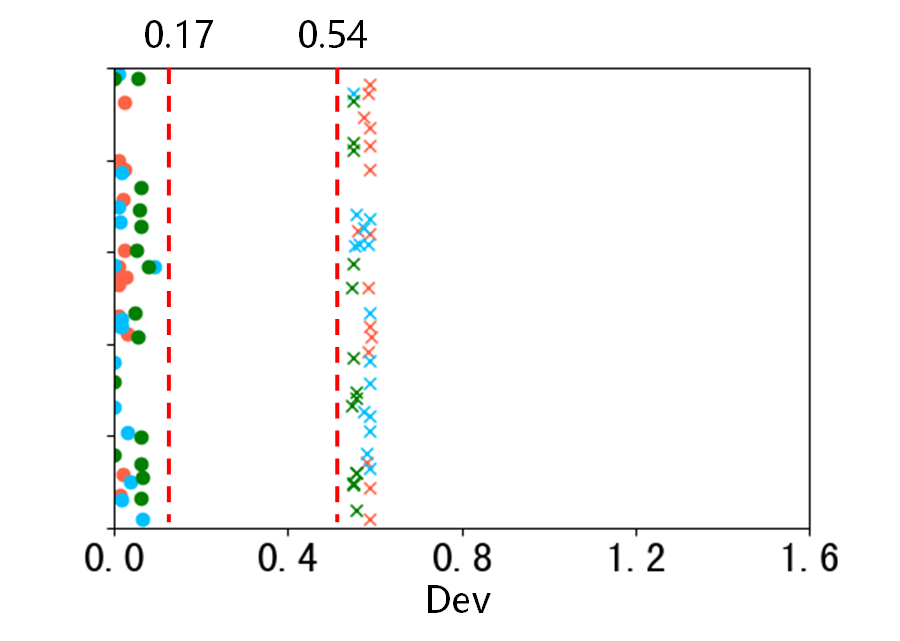}}
    \subfigure[\fm{Non-IID 50\%-attack}]{
    \includegraphics[width=0.35\linewidth]{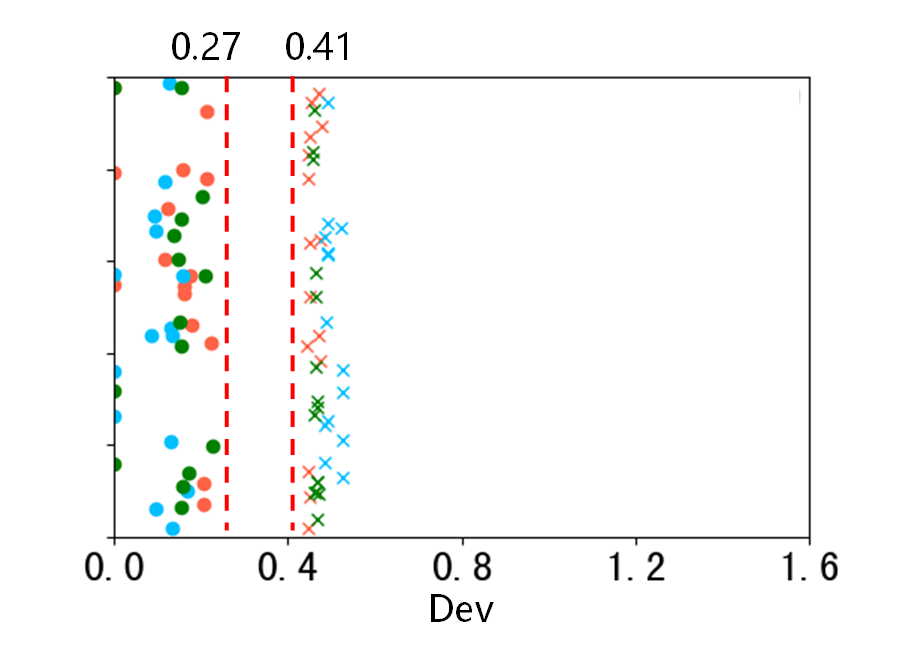}} \\
    \subfigure[\fm{IID 90\%-attack}]{
    \includegraphics[width=0.35\linewidth]{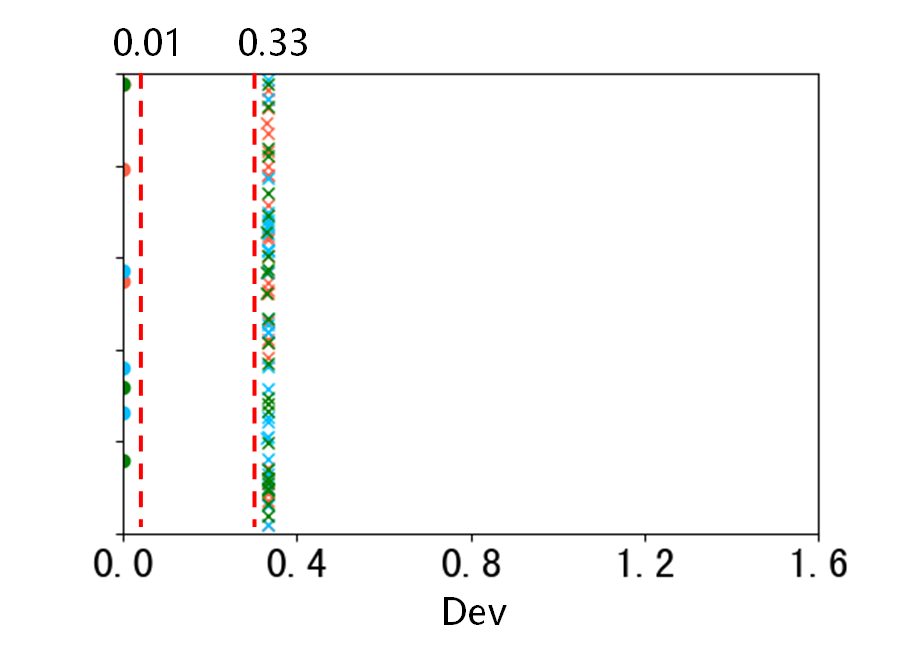}}
    \subfigure[\fm{Non-IID 90\%-attack}]{
    \includegraphics[width=0.35\linewidth]{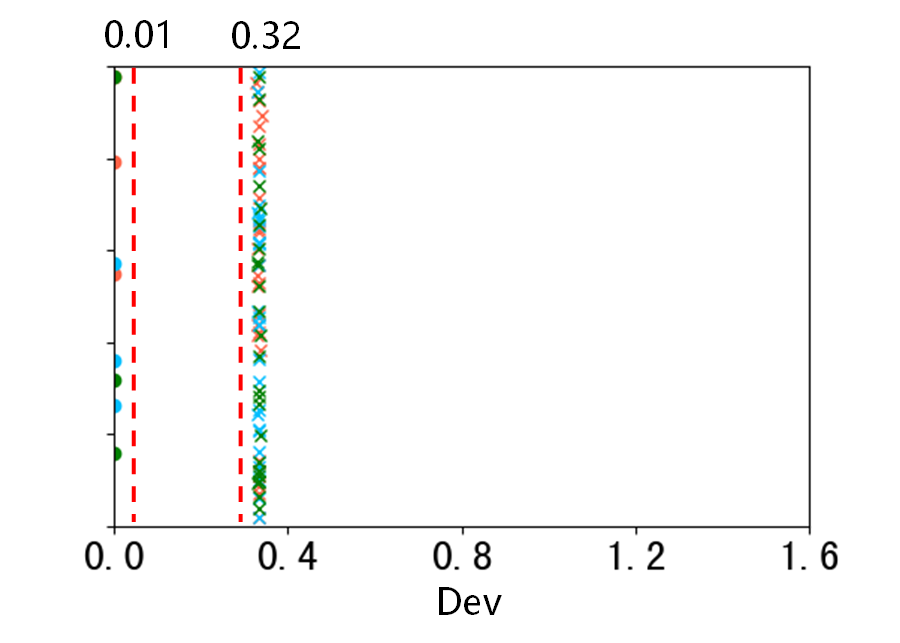}} \\    \includegraphics[width=0.7\linewidth]{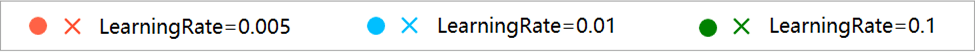}
   \caption{The reputation threshold boundary under different learning rates on MNIST dataset, 
   where ``$\bullet$'' and ``$\times$'' represent the benign client and the free-rider, respectively.}\label{mnist-3}
\end{figure}


\subsubsection{Hyperparameter Analysis of Reputation Threshold}
In this section, we investigate robust bounds on the reputation threshold.
The selection of reputation threshold separates free-riders from benign clients by grouping clients into $\left\{ P_n, P_r\right\}$.
A key challenge lies in choosing an appropriate reputation threshold, for example, a reputation threshold that is too large or too small may make it difficult to separate all free-riders from benign clients.

\textbf{Implementation Details.}
The similarity deviation values $Dev$ for all clients in the five datasets are tested and visualized under the IID and Non-IID settings, where $Dev$ takes the average of the first five rounds of clients.
Besides, we perform a unified analysis of the client proportions for the three free-rider attacks.
The result is shown in Fig.\ref{canshu10}.

\textbf{Results and Analysis.}
We through visual analysis, it is found that the reputation threshold selection of WEF-Defense has a certain boundary range, which explains why WEF-Defense can effectively separate benign clients and free-riders in various scenarios.
Fig.\ref{canshu10} shows that in the IID data scenario, the optional range of thresholds is larger than in the Non-IID data scenario. We guess that under the Non-IID data, there are some cases where the local data distribution of some benign clients is more extreme, resulting in a certain difference between its WEF-Matrix and other benign clients, but this does not affect the implementation of our method.
The reputation threshold set in the experiment can distinguish 100\% of benign clients and free-rider clients.

\subsubsection{Effect of Learning Rate on Reputation Threshold}
We analyze whether the learning rate has a strong effect on the bounds of the reputation threshold.

\textbf{Implementation Details.}
On the MNIST dataset under the IID and Non-IID settings, we consider the influence of different learning rates on the reputation threshold, where the learning rates are set to 0.005, 0.01, and 0.1, respectively.
The experimental results are shown in Fig.\ref{mnist-3}.

\textbf{Results and Analysis.}
The similarity deviation of the client does not fluctuate greatly under different learning rates, as can be seen from the analysis Fig.\ref{mnist-3},
indicating that the effect of the learning rate on the threshold boundary is small.
The reason may be that regardless of the setting of the learning rate, 
the optimization of the weights requires a variation process, 
which does not affect the formation of the WEF-Matrix.
It further demonstrates that the reputation threshold is not affected by the learning rate.

\begin{center}
\fcolorbox{black}{white!20}{\parbox{0.97\linewidth}
{
\emph{\textbf{Answer to RQ5:}}
Experiments demonstrate that WEF-Defense is robust to adaptive attacks and hyperparameter $\epsilon$.
Specifically,
1) due to the significant difference between benign clients and free-riders, 
WEF-Defense has a strong ability to resist camouflage and can effectively defend against adaptive attacks;
2) the hyperparameter $\epsilon$ in WEF-Defense has a good adjustable range, and is not greatly affected by the learning rate.
}
}
\end{center}

\section{Limitation and Discussion}\label{disccusion}
Although WEF-Defense has demonstrated its outstanding performance in defending against various free-rider attacks, 
its effectiveness can still be improved in terms of Non-IID data and time cost.

\textbf{Process Non-IID data.} 
The reputation threshold boundary range under the Non-IID setting is not as wide as that under the IID setting. 
We speculate the reason is that 
there are several benign clients with poor local data quality under the Non-IID setting.
These clients' contribution to federated training may not be much more than that of free-riders. 
Therefore, it is necessary to improve the identification of free-riders under the Non-IID setting.

\textbf{Reduce time cost.}
Despite the advantages of WEF-Defense in terms of defense, it can be further improved in terms of time cost.
The main reason is that the client needs to upload additional information, which increases the time cost. 
It is worth the effort to reduce the time cost while ensuring the defense effectiveness. 

\section{Conclusion}\label{conclusion}
In this paper, we highlight that the difference between free-riders and benign clients in the dynamic training progress can be effectively used to defend against free-rider attacks, 
based on which we propose WEF-Defense. 
WEF-Defense generally outperforms all baselines and also performs well against various camouflaged free-rider attacks. 
The experiments further analyze the effectiveness of WEF-Defense from five perspectives, 
and verify that WEF-Defense can not only defend against free-rider attacks, 
but also does not affect the training of benign clients. 
Since WEF-Defense and existing methods are complementary to each other,
we plan to design a more robust and secure federated learning mechanism by exploring the potential of combining them in the future work.
Besides, 
it is possible to conduct free-rider attack on vertical FL. 
In the future, 
we will explore the free-rider attack on vertical FL and possible defense.

\section{Acknowledgements}\label{sec14}
This research is supported by 
the National Natural Science Foundation of China (No. 62072406), 
the National Key R\&D Projects of China (No. 2018AAA0100801),
the Key R\&D Projects in Zhejiang Province (No. 2021C01117),
the 2020 Industrial Internet Innovation Development Project (No.TC200H01V), 
``Ten Thousand Talents Program'' in Zhejiang Province (No. 2020R52011), and 
the Key Lab of Ministry of Public Security (No. 2020DSJSYS001).


\newpage

\appendix
\section{More Visualizations\label{appendix_visualization}}

\begin{figure}[thb]
\centering
    \subfigure[bengin clients on MNIST dataset]{
    \includegraphics[width=0.95\linewidth]{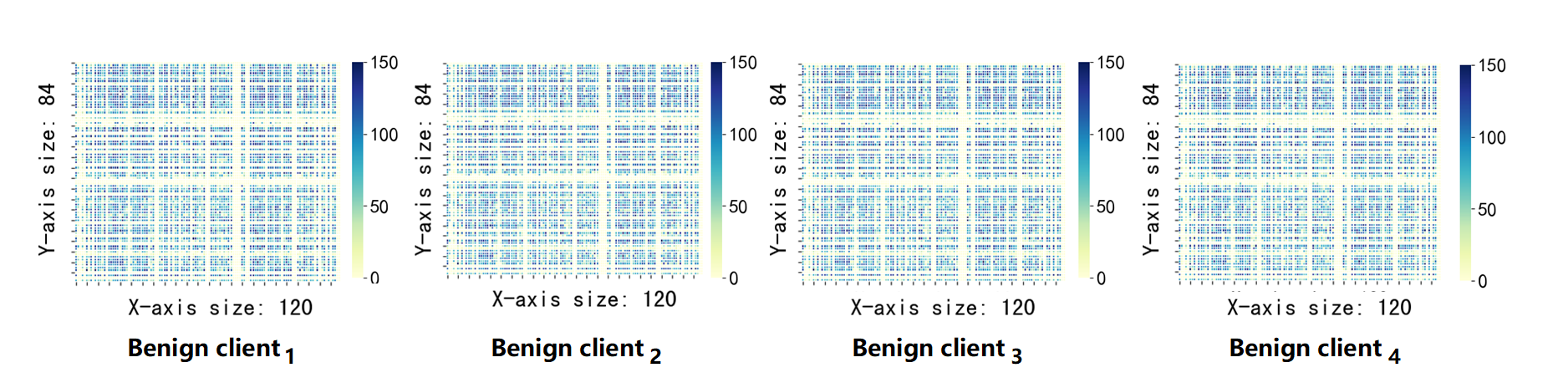}}\\
    \vspace{-2mm}
    \subfigure[\fm{ free-riders on MNIST dataset}]{
    \includegraphics[width=0.95\linewidth]{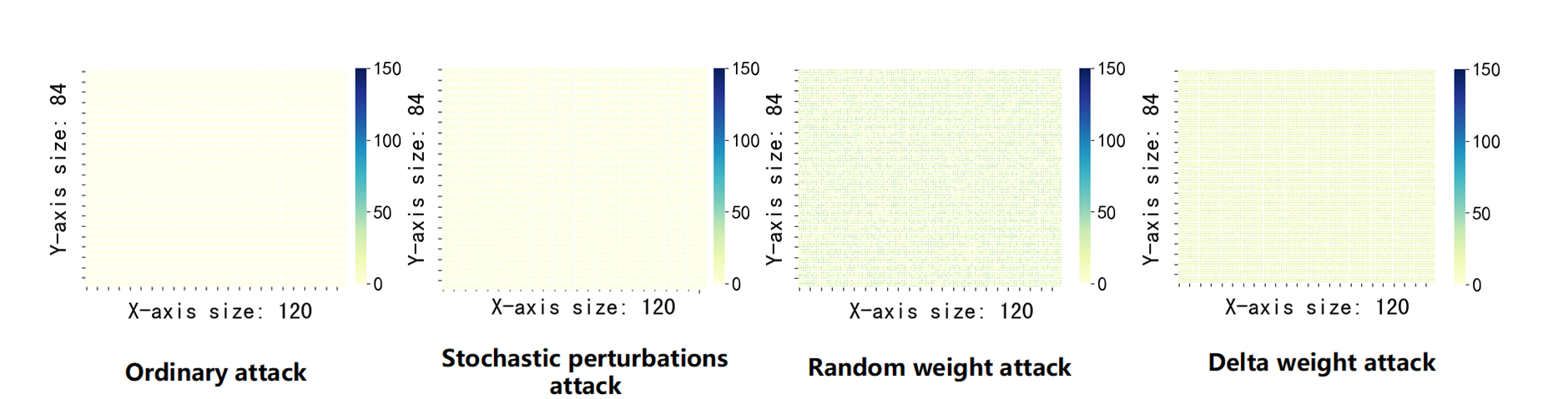}}\\
    \vspace{-2mm}
    \subfigure[ bengin client on BANK dataset]{
    \includegraphics[width=0.95\linewidth]{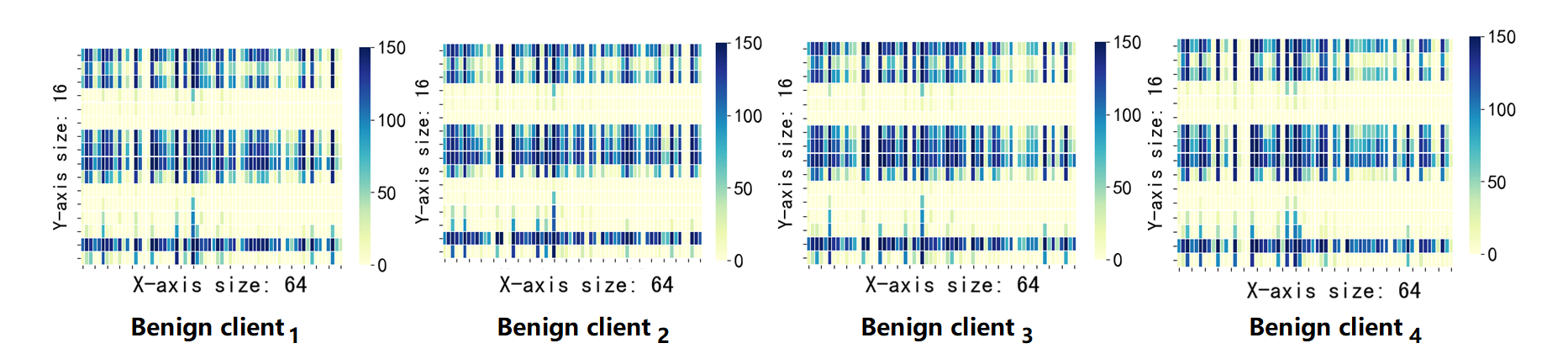}} \\
    \vspace{-2mm}
    \subfigure[\fm{ free-riders on BANK dataset}]{
    \includegraphics[width=0.95\linewidth]{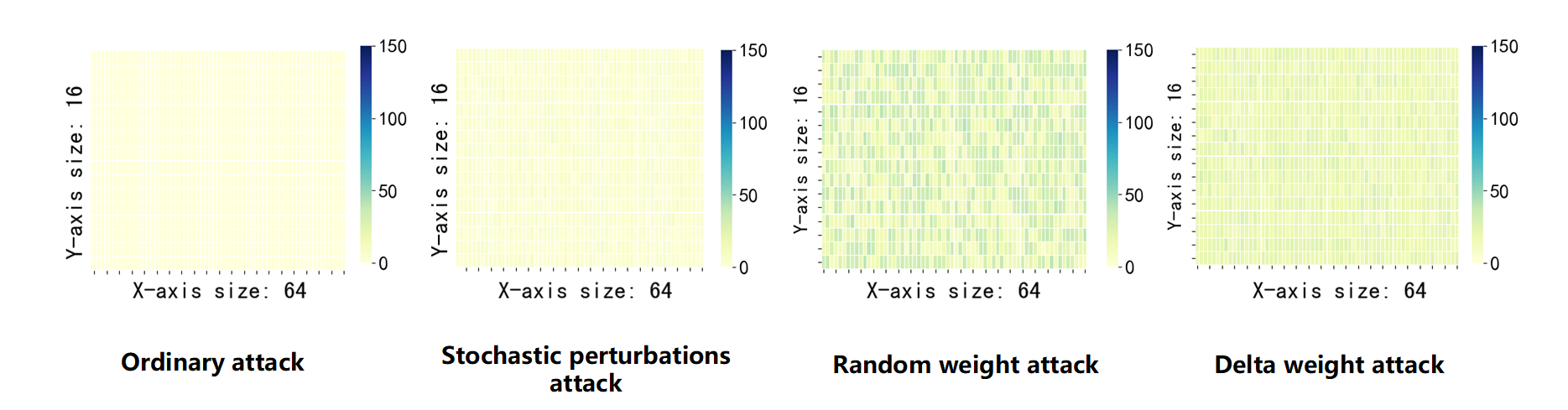}}
   \caption{Benign clients and free-riders' WEF-Matrix are visualized to observe their differences.}
\end{figure}

\begin{figure}[]
\centering
    \subfigure[The variation process of the global models' accuracy when the free-rider ratio is 10\%]{
    \includegraphics[width=0.3\linewidth]{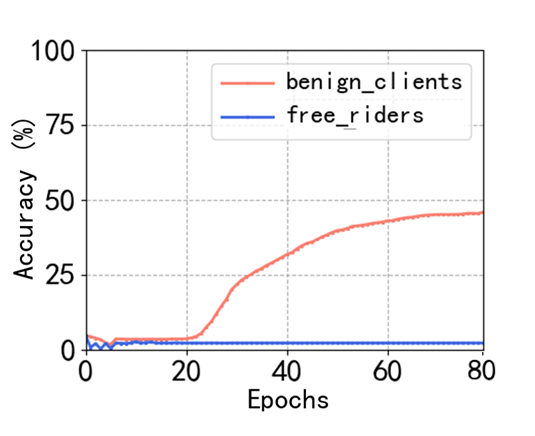}
    \includegraphics[width=0.3\linewidth]{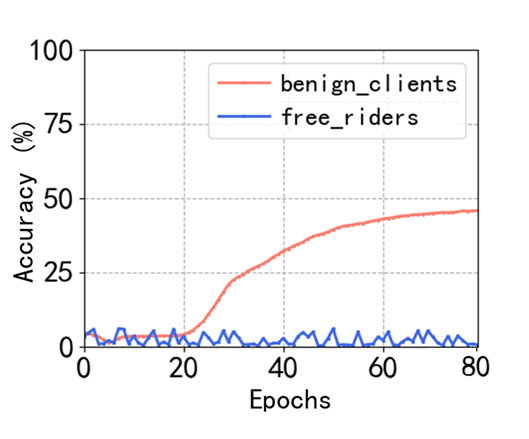}
    \includegraphics[width=0.3\linewidth]{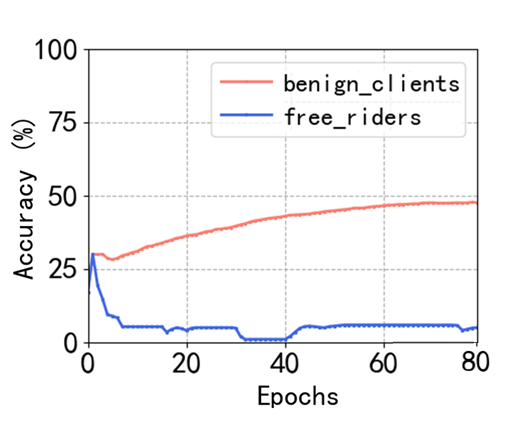}    } \\
     \subfigure[The variation process of the global models' accuracy when the free-rider ratio is 30\%]{
    \includegraphics[width=0.3\linewidth]{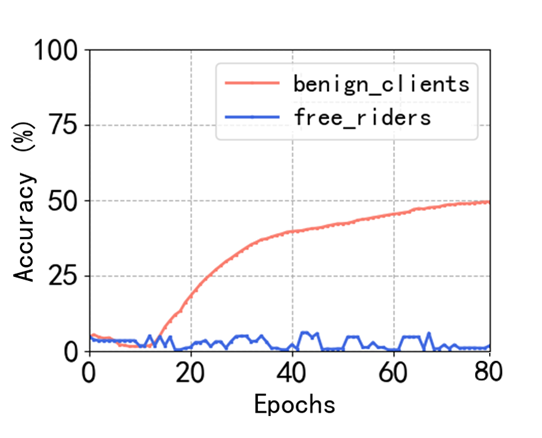}
    \includegraphics[width=0.3\linewidth]{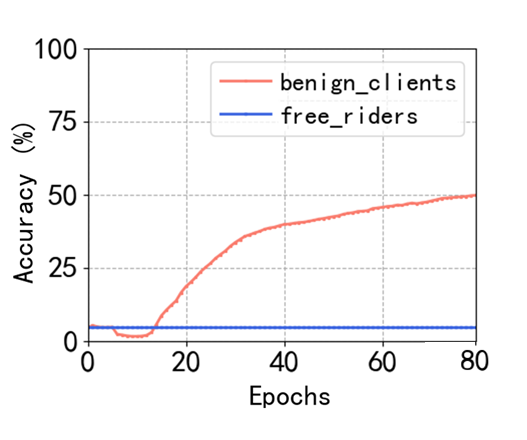}
    \includegraphics[width=0.3\linewidth]{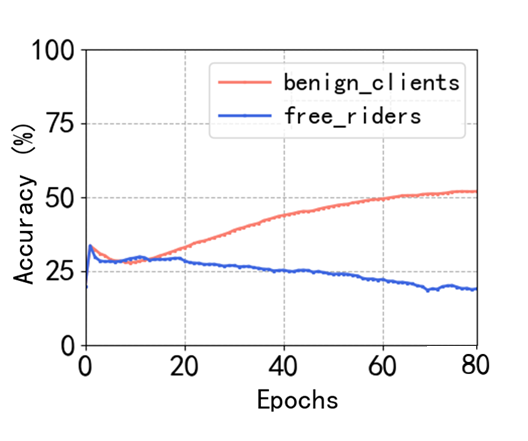}}\\
     \subfigure[The variation process of the global models' accuracy when the free-rider ratio is 50\%]{
    \includegraphics[width=0.3\linewidth]{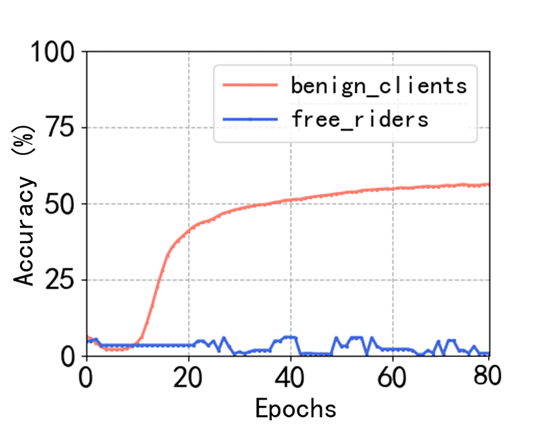}
    \includegraphics[width=0.3\linewidth]{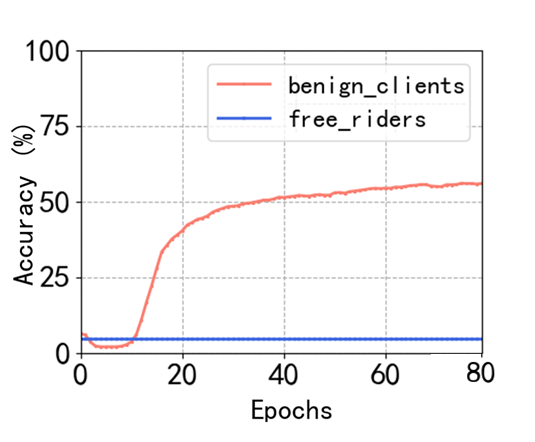}
    \includegraphics[width=0.3\linewidth]{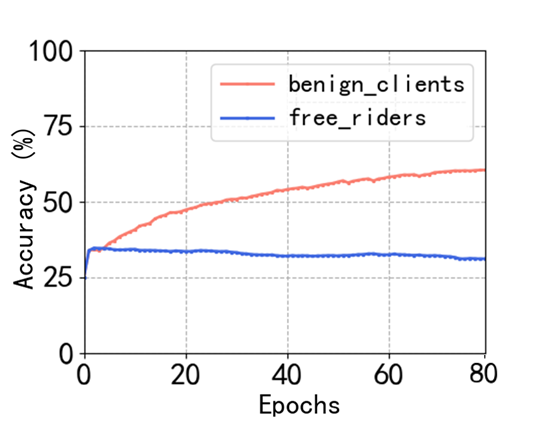}}\\
     \subfigure[The variation process of the global models' accuracy when the free-rider ratio is 90\%]{
    \includegraphics[width=0.3\linewidth]{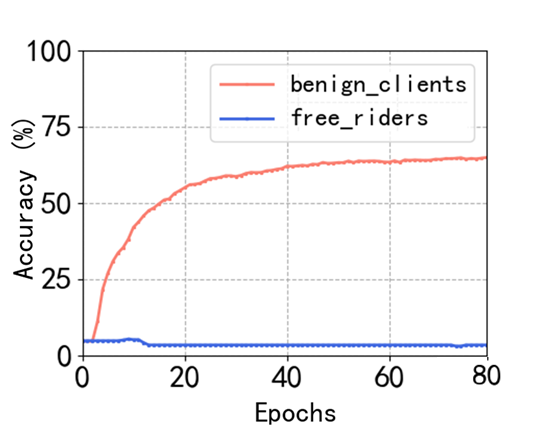}
    \includegraphics[width=0.3\linewidth]{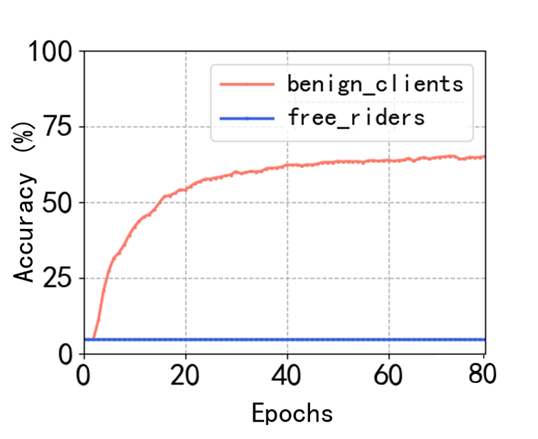}
    \includegraphics[width=0.3\linewidth]{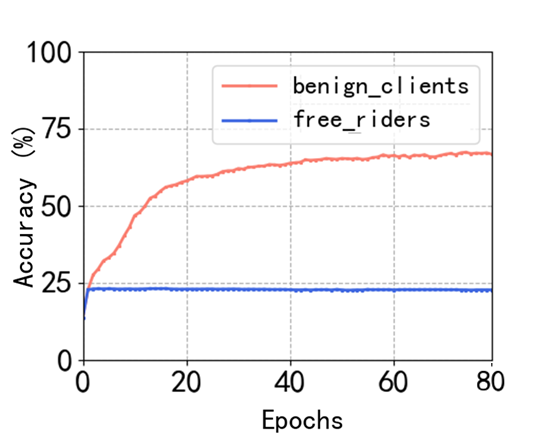}}
   \caption{
The process of global models' accuracy variation obtained by benign clients and free-riders during personalized federation training in the GTSRB dataset under the Non-IID setting.
For each subfigure, from left to right, experimental results of WEF-Defense against random weight attack, stochastic perturbations attack and delta weight attack are shown.
}
\end{figure}

\end{document}